\documentclass[11pt]{article}
\usepackage[a4paper, margin=1in]{geometry}
\usepackage[usenames,dvipsnames,svgnames,table]{xcolor} 

\usepackage[T1]{fontenc}
\usepackage{lmodern}
\usepackage[utf8]{inputenc}

\usepackage{algorithm}
\usepackage{algorithmic}
\floatname{algorithm}{Algorithm}

\usepackage{longtable}
\usepackage{hhline}
\usepackage{float}

\usepackage{xspace}
\usepackage{enumitem} 
\usepackage{rotfloat} 
\usepackage{graphicx}
\usepackage{amssymb, amsmath, amsthm, amscd}
\usepackage{tikz}
\usetikzlibrary{3d, calc}
\usepackage{verbatim}
\usepackage{natbib}
\usepackage{authblk}
\usepackage{multirow}
\usepackage{subcaption} 
\usepackage{array}
\newcolumntype{L}{>{\centering\arraybackslash}m{0.1\linewidth}}

\newtheorem{theorem}{Theorem}[section]
\newtheorem{lemma}[theorem]{Lemma}
\newtheorem{proposition}[theorem]{Proposition}
\newtheorem{corollary}[theorem]{Corollary}

\usepackage{hyperref}

\newcommand{\bbR}{\mathbb{R}}

\newcommand{\calF}{\mathcal{F}}

\newcommand{\calN}{\mathcal{N}}

\newcommand{\calP}{\mathcal{P}}

\newcommand{\barmu}{\bar{\mu}}

\newcommand{\Frechet}{Fr\'{e}chet\xspace}

\begin{document}
	
\title{A Particle-Flow Algorithm for Free-Support Wasserstein Barycenters}
\author[1]{Kisung You}
\affil[1]{Department of Mathematics, Baruch College}
\date{}

\maketitle
\begin{abstract}
The Wasserstein barycenter extends the Euclidean mean to the space of probability measures by minimizing the weighted sum of squared 2-Wasserstein distances. We develop a free-support algorithm for computing Wasserstein barycenters that avoids entropic regularization and instead follows the formal Riemannian geometry of Wasserstein space. In our approach, barycenter atoms evolve as particles advected by averaged optimal-transport displacements, with barycentric projections of optimal transport plans used in place of Monge maps when the latter do not exist. This yields a geometry-aware particle-flow update that preserves sharp features of the Wasserstein barycenter while remaining computationally tractable. We establish theoretical guarantees, including consistency of barycentric projections, monotone descent and convergence to stationary points, stability with respect to perturbations of the inputs, and resolution consistency as the number of atoms increases. Empirical studies on averaging probability distributions, Bayesian posterior aggregation, image prototypes and classification, and large-scale clustering demonstrate accuracy and scalability of the proposed particle-flow approach, positioning it as a principled alternative to both linear programming and regularized solvers.
\end{abstract}

\section{Introduction}

Optimal transport (OT) is an emerging branch of mathematical science that studies the geometry of the space of probability measures and has rapidly become a cornerstone in modern data analysis. Originating from Monge's eighteenth-century formulation \citep{monge_1781_MemoireTheorieDeblais} and rigorously generalized by Kantorovich in the mid-twentieth century \citep{kantorovitch_1958_TranslocationMasses}, OT provides a principled way of quantifying discrepancy between probability measures by explicitly modeling the mass displacement required to transform one into another. Beyond its intrinsic geometric and analytic appeal, OT has found an extraordinary breadth of applications across disciplines: in statistics for parameter estimation \citep{bernton_2019_ParameterEstimationWasserstein} and hypothesis testing \citep{ramdas_2017_WassersteinTwoSampleTesting}, in machine learning for generative adversarial networks \citep{arjovsky_2017_WassersteinGenerativeAdversarial} and domain adaptation \citep{courty_2017_OptimalTransportDomain}, and in computational biology for modeling trajectories of high-dimensional single-cell distributions \citep{schiebinger_2019_OptimalTransportAnalysisSingleCell}, to name a few. Its success lies in its ability to geometrize the space of probability measures, capturing global structure that goes beyond classical metrics such as Kullback-Leibler divergence or total variation, and enabling new approaches for inference and learning from complex data.

A central problem within this framework is the Wasserstein barycenter, which extends the Euclidean notion of a mean to the space of probability measures under the Wasserstein metric. Just as the arithmetic mean minimizes the sum of squared Euclidean distances to a set of points, the Wasserstein barycenter minimizes the weighted sum of squared Wasserstein distances to a collection of distributions \citep{agueh_2011_BarycentersWassersteinSpace}. This seemingly simple construction has deep consequences, providing a canonical way of averaging distributions that may live on different supports or exhibit non-Euclidean geometry. Applications of barycenters span a wide spectrum. In statistics, they appear in consensus learning and Bayesian analysis, where multiple posterior distributions are aggregated across data sources \citep{srivastava_2018_ScalableBayesBarycenter}. In computer vision, barycenters enable image morphing, texture synthesis, and shape interpolation \citep{solomon_2015_ConvolutionalWassersteinDistances}. In data science more broadly, barycenters serve as natural cluster representatives in distribution $k$-means where each cluster of probability measures is summarized by its Wasserstein barycenter as a representative centroid \citep{zhuang_2022_WassersteinKmeansClustering}.

On the computational side, however, the barycenter problem is notoriously demanding, especially in the setting of empirical measures where input distributions are represented as point clouds. The classical approach formulates barycenter estimation as a large-scale linear program (LP) over couplings between supports \citep{bazaraa_2010_LinearProgrammingNetwork}. While exact, the number of variables grows quadratically in the support size of the inputs and linearly in the number of measures, rendering the approach impractical for modern datasets. To overcome this bottleneck, a significant body of work has shifted toward regularized approximations, most prominently entropic regularization combined with Sinkhorn iterations \citep{cuturi_2013_SinkhornDistancesLightspeed, benamou_2015_IterativeBregmanProjections}. These methods leverage the favorable structure of entropic OT, where couplings can be computed by iterative matrix scaling in which GPU acceleration is straightforward, and stochastic variants allow for streaming and large-scale applications \citep{genevay_2016_StochasticOptimizationLargescale}. Yet, this computational efficiency comes at a cost in that regularization introduces a bias whose effect depends on the chosen parameter and the smoothing effect can blur sharp features of the barycenter, limiting interpretability in applications that rely on precise structural preservation. As such, the landscape of barycenter computation is marked by a fundamental trade-off between exact but intractable LP formulations on one side and scalable but biased regularized solvers on the other.

In this paper, we propose a different route, grounded in the formal Riemannian geometry of the Wasserstein space \citep{otto_2001_GeometryDissipativeEvolution}, that aims at directly solving the problem without regularization. We view barycenter computation as an optimization problem on a curved manifold of probability measures, where the objective functional acts as an energy landscape. Otto's calculus shows that the Wasserstein gradient flow corresponds to transporting mass along displacement fields defined by OT maps, which naturally yields a particle-flow interpretation in the empirical setting. To briefly describe, when the barycenter is represented by a discrete measure with free support, each atom behaves as a particle that is advected along an averaged OT displacement toward the input measures. To make this tractable, we approximate the maps with barycentric projections of OT plans. The resulting algorithm can be seen as an explicit Euler discretization of the Wasserstein gradient flow, producing an unregularized, geometry-aware flow that preserves sharp features of the barycenter while remaining computationally tractable. The theoretical part of the paper establishes approximation guarantees for barycentric projections, convergence of the proposed free-support scheme, and stability with respect to perturbations of the inputs. On the practical side, we show that the algorithm is efficient to implement, enjoys monotonic descent, and performs well in simulations and real data examples. Altogether, our contributions position the method as a principled particle-flow alternative to both LP-based and regularized barycenter solvers, balancing rigor with computational feasibility.  

We are not the first to view the barycenter problem through the lens of \Frechet means and to employ gradient descent methods, as in \citet{zemel_2019_FrechetMeansProcrustes}. However, our approach differs in that we focus on the free-support empirical barycenter, systematically replacing Monge maps that are often unavailable with barycentric projections of optimal transport plans, avoiding entropic regularization, and providing new guarantees that are tailored to the discrete and large-scale settings most relevant in modern applications.

The remainder of the paper is organized as follows. 
Section~\ref{sec:chp2-prelim} reviews the geometry of the Wasserstein space of order 2, Otto's formal Riemannian framework, and the barycenter stationarity condition. Then, we develop our free-support algorithm in Section~\ref{sec:chp3-methods} by deriving a Wasserstein gradient descent for the barycenter objective, replacing unavailable Monge maps by barycentric projections of optimal plans, and giving an implementable particle-flow update with practical guidance on algorithmic components. Section~\ref{sec:chp4-theory} presents theoretical guarantees including consistency of barycentric projections, monotone descent and convergence to stationary points, stability with respect to perturbations in the inputs, and resolution consistency as the number of barycenter atoms increases. 
Section~\ref{sec:chp5-experiment} reports empirical results using simulated and real data examples on averaging probability distributions, Bayesian posterior aggregation, image prototyping and classification, and large-scale clustering with distributed vector quantization. We conclude in Section~\ref{sec:chp6-conclusion} with a summary and several avenues for future work. The proposed algorithm is implemented in an \textsf{R} package \textsf{T4transport} \citep{you_2020_T4transportToolsComputational} available at \url{https://kisungyou.com/T4transport}.

\section{Preliminaries}\label{sec:chp2-prelim}

In this section, we introduce the mathematical framework that supports our algorithmic development. We begin by reviewing the geometry of Wasserstein space, focusing on the distance and geodesic structure, followed by presentation of Otto's Riemannian framework, which equips the space with a formal tangent and inner product structure \citep{otto_2001_GeometryDissipativeEvolution, ambrosio_2005_GradientFlowsMetric}. Finally, we introduce the Wasserstein barycenter and its characterization via a tangent-space stationarity condition.

\subsection{The Wasserstein space and displacement interpolation}

The Wasserstein distance arises from the theory of optimal transport, which seeks to quantify the cost of transforming one probability distribution into another. Historically, this problem was first posed by Gaspard Monge in 1781, and later relaxed and rigorously reformulated by Leonid Kantorovich in the 20th century. We introduce the Kantorovich formulation first, which is more general and foundational to the geometry of the Wasserstein space.

Let $\calP_2(\bbR^d)$ denote the space of Borel probability measures on $\bbR^d$ with finite second moments:
\begin{equation*}
    \calP_2(\bbR^d) := \left\lbrace \mu \in \calP(\bbR^d) \bigg\vert \int_{\bbR^d} \|x\|^2 d\mu(x) < \infty\right\rbrace.
\end{equation*}
The Kantorovich optimal transport problem defines the squared Wasserstein distance between $\mu,\nu \in \calP_2(\bbR^d)$ as
\begin{equation*}
    W_2^2 (\mu, \nu)  := 
    \underset{\gamma \in \Pi(\mu,\nu)}{\inf}~ \int_{\bbR^d \times \bbR^d} \|x-y\|^2 d\gamma(x,y)
    ,
\end{equation*}
where $\Pi(\mu,\nu)$ denotes the set of all joint distributions on $\bbR^d \times \bbR^d$, also called couplings between $\mu$ and $\nu$, with marginals $\mu$ and $\nu$. This formulation allows for mass splitting that a single point $x$ under $\mu$ may be split and sent to multiple destinations under $\nu$ through $\gamma$. This relaxation guarantees the existence of a solution under mild conditions.

The original transport problem, proposed by \cite{monge_1781_MemoireTheorieDeblais}, aims at finding a deterministic map $T:\bbR^d \to \bbR^d$ that pushes $\mu$ forward to $\nu$, i.e., $\mu(T^{-1}(A)) = \nu(A)$ for all Borel measurable $A \subset \bbR^d$, and minimizes the transport cost 
\begin{equation*}
    \inf_{T{\#} \mu = \nu}~ \int_{\bbR^d} \|x-T(x)\|^2 d\mu(x).
\end{equation*}
This is a much more constrained formulation. Unlike Kantorovich's, Monge's formulation does not allow splitting of mass. Moreover, it is highly non-convex and may not even admit a solution. Nevertheless, when $\mu$ is absolutely continuous with respect to the Lebesgue measure, the Kantorovich problem is known to admit a unique optimal transport map $T_{\mu \to \nu}$ and the Monge and Kantorovich formulations coincide. 

Equipped with this metric structure, $\calP_2(\bbR^d)$ becomes a geodesic space \citep{burago_2001_CourseMetricGeometry}. Between any two distributions $\mu_0, \mu_1 \in \calP_2(\bbR^d)$, the shortest path is given by displacement interpolation \citep{mccann_1997_ConvexityPrincipleInteracting}
\begin{equation*}
    \mu_t := ((1-t)\text{Id} + tT)\#\mu_0,\quad t\in [0,1],
\end{equation*}
where $T$ is the optimal transport map from $\mu_0$ to $\mu_1$ and $\text{Id}$ is the identity map. This interpolation reflects the physical movement of mass and ensures that intermediate distributions lie along geodesics under the $W_2$ metric.

\subsection{First-order geometry of the Wasserstein space}
\cite{otto_2001_GeometryDissipativeEvolution} introduced a formal Riemannian viewpoint on $\calP_2(\bbR^d)$, interpreting it as an infinite-dimensional Riemannian manifold in which probability distributions evolve along gradient flows \citep{ambrosio_2005_GradientFlowsMetric}. Although this structure does not satisfy all properties of classical differential geometry, it provides a powerful first-order calculus for defining tangent spaces, inner products, and components for numerical optimization on the space of probability measures. 

Let $(\mu_t)_{t\in [0,1]}$ be a smooth path of probability measures. Its infinitesimal evolution is governed by the continuity equation
\begin{equation*}
    \frac{d}{dt}\mu_t + \nabla \cdot (\mu_t v_t) = 0,
\end{equation*}
where $v_t:\bbR^d \to \bbR^d$ is a time-dependent velocity field describing the motion of probability mass through space. From this dynamical perspective, the tangent space at $\mu \in \calP_2(\bbR^d)$, denoted as $T_\mu \calP_2(\bbR^d)$, consists of velocity fields that are square-integrable under $\mu$, and arises from gradients of smooth scalar functions. To be precise, it is given by 
\begin{equation*}
    T_\mu \calP_2(\bbR^d) := \overline{\left\lbrace -\nabla \phi:\phi \in C_c^\infty(\bbR^d) \right\rbrace}^{L^2(\mu)},
\end{equation*}
where $C_c^\infty(\bbR^d)$ denotes the space of smooth functions with compact support. The minus gradient field $-\nabla \phi$ represents a velocity direction along which the measure $\mu$ is infinitesimally transported according to the continuity equation. These gradients serve as infinitesimal generators of mass-preserving flows and are intimately connected to optimal transport theory, where displacement maps between measures arise as gradients of convex potentials. The use of compactly supported smooth functions ensures a dense subset for approximation, while the closure in the $L^2(\mu)$ norm guarantees that the tangent space includes all square-integrable vector fields that can be approximated in this manner. This construction ensures that $T_\mu \calP_2(\bbR^d)$ is a Hilbert space and aligns with the formal Riemannian geometry underlying the Wasserstein space.

This tangent space is equipped with the natural inner product
\begin{equation*}
    \langle v,w\rangle_{T_\mu} := \int_{\bbR^d} \langle v(x), w(x) \rangle d\mu(x),
\end{equation*}
which turns $T_\mu \calP_2(\bbR^d)$ into a Hilbert space, allowing notions of distance, angle, and projection to be well defined. Similarly, for $\mu, \nu \in \calP_2(\bbR^d)$ and $v \in T_\mu \calP_2(\bbR^d)$,  the logarithmic map $\log_\mu:\calP_2(\bbR^d) \to T_\mu \calP_2(\bbR^d)$ and the exponential map $\exp_\mu:T_\mu \calP_2(\bbR^d) \to \calP_2(\bbR^d)$ are defined as follows:
\begin{equation*}
    \log_\mu(\nu) := T_{\mu\to\nu} - \text{Id},\qquad \exp_\mu(v) := (\text{Id}+v){\#} \mu,
\end{equation*}
which enable one to define geodesics, compute gradients, and construct optimization schemes directly over $\calP_2(\bbR^d)$, mirroring analogous constructions on finite-dimensional Riemannian manifolds.

Despite its utility, the Riemannian structure is formal. For the rest of the paper, we abuse the notation $\log_\mu(\nu)=T_{\mu\to\nu}-\mathrm{Id}$ and
$\exp_\mu(v)=(\mathrm{Id}+v)_{\#}\mu$ in this sense. The space $\calP_2(\bbR^d)$ does not possess coordinate charts on a smooth manifold atlas, as tangent vectors are defined indirectly through optimal transport rather than local linearization. When $\mu$ is absolutely continuous and $T_{\mu\to\nu}$ is the unique optimal map, the curve $t\mapsto ((1-t)\text{Id} + tT_{\mu\to\nu})_{\#}\mu$ is the $W_2$-geodesic from $\mu$ to $\nu$. Particularly, if $v = T_{\mu\to \nu}-\text{Id}$, then $\exp_\mu(v)$ is the geodesic endpoint at $t=1$. For an arbitrary $v\in T_\mu\mathcal P_2$, the quantity $(\text{Id}+v)_{\#}\mu$ needs not lie on a geodesic, hence we use it only as a first-order Eulerian step within Otto's formal calculus. In general, key differential geometric tools such as the Levi-Civita connection, curvature tensor, and parallel transport are not available. Moreover, the exponential map may be discontinuous or fail to be invertible, particularly for measures with discrete or singular support.

\subsection{Wasserstein barycenter}

Let $\mu_1, \ldots, \mu_N \in \calP_2(\bbR^d)$ be a collection of probability measures with weights $(\pi_1, \ldots, \pi_N) \in \mathring{\Delta}^{N-1}$ in the probability simplex, i.e., $\pi_n > 0$ for all $n\in [N]:=\lbrace 1, 2, \ldots, N\rbrace$ and $\sum_{n=1}^N \pi_n = 1$. The Wasserstein barycenter \citep{agueh_2011_BarycentersWassersteinSpace}
of this sample is defined as the minimizer of the weighted functional
\begin{equation}\label{eq:definition_barycenter}
    \bar{\mu} := \underset{\mu \in \calP_2(\bbR^d)}{\arg\min}~ \sum_{n=1}^N \pi_n W_2^2(\mu, \mu_n).
\end{equation}
This is a direct generalization of the mean from Euclidean space to the space of probability measures by exploiting the characterization of the sample mean as a minimizer of the sum of squared distances. Although its existence is readily guaranteed under mild conditions that all $\mu_n$'s are in $\calP_2(\bbR^d)$, the uniqueness of the Wasserstein barycenter typically requires additional regularity conditions, such as when at least one of $\mu_n$'s is absolutely continuous.

From Otto's formal Riemannian structure of the Wasserstein space, the barycenter satisfies the following stationarity condition \citep{lanzetti_2025_FirstOrderConditionsOptimization}:
\begin{equation}\label{eq:stationarity_condition}
    \sum_{n=1}^N \pi_n \log_{\bar{\mu}}(\mu_n) = 0 \in T_{\bar{\mu}}\calP_2(\bbR^d).
\end{equation}
This condition mirrors the first-order condition for the Euclidean mean, stating that the weighted average of the displacement fields from $\bar{\mu}$ to each $\mu_n$ as defined by the optimal transport maps $\log_{\bar{\mu}}(\mu_n) = T_{\bar{\mu} \to \mu_n} - \text{Id}$ must vanish. That is, the optimal transport directions from $\bar{\mu}$ to the input distributions are in equilibrium. From a computational point of view, this stationarity condition claims that one may search for a distribution $\mu \in \calP_2(\bbR^d)$ for which the average log map vanishes.

\section{Methods}\label{sec:chp3-methods}

\subsection{Revisiting the gradient descent in the Wasserstein space}

We now revisit the Wasserstein barycenter problem as a Riemannian optimization problem over the space $\calP_2(\bbR^d)$. In Otto's formal calculus, this space is endowed with a Riemannian-like structure in which the squared Wasserstein distance induces a first-order differential geometry. This requires characterizing the gradient of the squared Wasserstein distance functional under the formal Riemannian structure. Specifically, we consider the map 
\begin{equation}
    \calF_\nu(\mu) = W_2^2(\mu,\nu),
\end{equation}
for $\mu, \nu \in \calP_2(\bbR^d)$. \cite{zemel_2019_FrechetMeansProcrustes} showed that the Riemannian gradient of $\calF_\nu(\mu)$ is $-2\log_{\mu}(\nu)$ given by the logarithmic map. In the following lemma, we reiterate this explicit construction from the formal geometric perspective.

\begin{lemma}
    Let $\mu,\nu \in \calP_2(\bbR^d)$ with $\mu$ absolutely continuous with respect to the Lebesgue measure and let $T = \nabla \psi$ be the unique optimal transport map from $\mu$ to $\nu$, assumed to exist and be unique, for a convex function $\psi:\bbR^d \to \bbR$. The Riemannian gradient of the functional $\calF_\nu(\mu) = W_2^2(\mu,\nu)$ under the formal Riemannian structure on $\calP_2(\bbR^d)$ is given by 
    \begin{equation}\label{eq:gradient_of_squared_distance}
        \text{grad}_\mu ~\calF_\nu(\mu) = -2\log_\mu(\nu),
    \end{equation}
    where the logarithmic map is defined pointwise by $\log_\mu(\nu)(x) = T(x)-x$.
\end{lemma}
\begin{proof}
We begin by considering a smooth perturbation of $\mu$ along a direction 
$-\nabla \phi \in T_\mu \mathcal P_2(\mathbb R^d)$. Define 
$\mu_\epsilon := (\mathrm{Id}-\epsilon \nabla\phi)_{\#}\mu$, which is a
smooth curve in $\mathcal P_2(\mathbb R^d)$ for small $\epsilon>0$.  
By the differentiability of the squared Wasserstein distance along such
perturbations \citep{ambrosio_2005_GradientFlowsMetric}, the first
variation of $\mathcal F_\nu(\mu_\epsilon)$ in $\epsilon$ can be computed by
pulling back the integral against $\mu$. Writing $y = x - \epsilon \nabla
\phi(x)$, we have $d\mu_\epsilon(y) = d\mu(x)$ up to $o(\epsilon)$, while the
optimal map $S_\epsilon$ from $\mu_\epsilon$ to $\nu$ satisfies
$S_\epsilon(y) \approx T(x)$ for $\epsilon \to 0$, where
$T=\nabla\psi$ is the Brenier map from $\mu$ to $\nu$. Hence
\[
\mathcal F_\nu(\mu_\epsilon) 
  = \int_{\mathbb R^d} \|T(x) - (x - \epsilon \nabla\phi(x))\|^2 \, d\mu(x) + o(\epsilon).
\]
Expanding the integrand yields
\[
\|T(x) - (x - \epsilon \nabla \phi(x))\|^2 
  = \|T(x) - x\|^2 + 2\epsilon \langle T(x)-x, \nabla \phi(x)\rangle + o(\epsilon).
\]
Substituting back, we obtain
\[
\mathcal F_\nu(\mu_\epsilon) 
  = \int_{\mathbb R^d} \|T(x)-x\|^2 d\mu(x) 
    + 2\epsilon \int_{\mathbb R^d} \langle T(x)-x, \nabla\phi(x)\rangle d\mu(x)
    + o(\epsilon).
\]
Therefore, the directional derivative at $\mu$ in the direction $-\nabla \phi$ is
\[
\frac{d}{d\epsilon}\mathcal F_\nu(\mu_\epsilon)\Big|_{\epsilon=0}
   = 2 \int_{\mathbb R^d} \langle T(x)-x, \nabla\phi(x)\rangle d\mu(x).
\]
By the definition of the Riemannian gradient in Otto’s formal geometry,
\[
\frac{d}{d\epsilon}\mathcal F(\mu_\epsilon)\Big|_{\epsilon=0}
  = \langle \mathrm{grad}_\mu \mathcal F, -\nabla\phi \rangle_{L^2(\mu)}
  = - \int_{\mathbb R^d} \langle \mathrm{grad}_\mu \mathcal F(x), \nabla\phi(x)\rangle d\mu(x).
\]
Since this equality holds for all $\phi \in C_c^\infty(\mathbb R^d)$, the
Riesz representation theorem \citep{stein_2011_FunctionalAnalysisIntroduction}
implies
\[
\mathrm{grad}_\mu \mathcal F_\nu(x) = -2\,(T(x)-x) = -2\,\log_\mu(\nu)(x),
\]
which completes the proof.
\end{proof}

This lemma sets the ground for an iterative optimization scheme for the objective functional $\calF(\mu) = \sum_{n=1}^N \pi_n W_2^2(\mu, \mu_n)$ as in Equation  \eqref{eq:definition_barycenter}. Using linearity of the gradient operator, the gradient of $\calF$ at a fixed $\mu$ is given by 
\begin{equation*}
    \text{grad}_\mu\calF = \sum_{n=1}^N \pi_n   \text{grad}_\mu W_2^2(\mu, \mu_n) = -2\sum_{n=1}^N \pi_n \log_\mu (\mu_n).
\end{equation*}
Hence, a natural Riemannian gradient descent proceeds by iteratively transporting the current iterate $\mu^{(k)}$ along the direction of steepest descent defined by $-\text{grad}_\mu \calF$, which can be done by the exponential map as follows:
\begin{equation}\label{eq:updating_log_exp}
    \mu^{(k+1)} = \exp_{\mu^{(k)}}\left(-\eta_k \cdot \text{grad}_{\mu^{(k)}}\calF\right) = \exp_{\mu^{(k)}}\left(2\eta_k \sum_{n=1}^N \pi_n \log_{\mu^{(k)}}(\mu_n)\right)
\end{equation}
where $\eta_k > 0$ denotes the step size. When $\mu^{(k)}$ is absolutely continuous, the updating rule \eqref{eq:updating_log_exp} can be written in terms of the transport maps
\begin{equation}\label{eq:updating_map}
\begin{aligned}
    \mu^{(k+1)} &= \exp_{\mu^{(k)}}\left(2\eta_k \sum_{n=1}^N \pi_n (T_{n}^{(k)} -\text{Id})\right) \\
      &= \left((1-2\eta_k)\text{Id} + 2\eta_k \sum_{n=1}^N \pi_n T_n^{(k)}\right){\#} \mu^{(k)} 
\end{aligned}
\end{equation}
where $T_n^{(k)}$ is the optimal transport map from $\mu^{(k)}$ to $\mu_n$. This means that at the $k$-th iteration we compute the convex combination of the optimal transport maps from $\mu^{(k)}$ to each $\mu_n$, apply the resulting map to $\mu^{(k)}$, and push it forward to obtain the next iterate.

\cite{zemel_2019_FrechetMeansProcrustes} investigated several properties of the gradient descent algorithm as described in \eqref{eq:updating_log_exp} and \eqref{eq:updating_map}. Suppose an iterate $\mu^{(k)}$ is absolutely continuous. Then, using the step size $\eta_k \in  [0,1/2]$ guarantees the absolute continuity of an update $\mu^{(k+1)}$. Furthermore, the optimal step size is analytically available at $\eta_k = 1/2$. It was also shown that if all $\mu_n$'s are absolutely continuous with at least one having bounded density, then the unique limit point of the sequence generated by the gradient descent is the unique \Frechet mean, i.e., the unique Wasserstein barycenter. 

\subsection{Free-support Wasserstein barycenter with empirical measures}
\subsubsection{Problem setup}
We now specialize the barycenter problem to the setting where all input measures $\mu_1, \ldots, \mu_N$ are empirical distributions with discrete support and potentially non-uniform weights. Each measure is represented as 
\begin{equation*}
    \mu_n = \sum_{j=1}^{m_n} w_{n,j} \delta_{x_{n,j}},\quad w_n = (w_{n,1}, \ldots, w_{n,m_n}) \in \mathring{\Delta}^{m_n-1},
\end{equation*}
where $\lbrace x_{n,j}\rbrace_{j=1}^{m_n} \subset \bbR^d$ denote the support points. Similarly, we consider a target Wasserstein barycenter of the form
\begin{equation*}
    \bar{\mu} = \sum_{i=1}^m v_i \delta_{z_i},\quad v=(v_1, \ldots, v_m) \in \mathring{\Delta}^{m-1},
\end{equation*}
where the support points $\lbrace z_i\rbrace_{i=1}^m \subset \bbR^d$ are free to be optimized given fixed weights $v$. A convenient yet common assumption is to set $v_i = 1/m$ with uniform weights. Then, the optimization problem becomes 
\begin{equation}
    \min_{\bar{\mu}}  \sum_{n=1}^N \pi_n W_2^2(\bar{\mu}, \mu_n),
\end{equation}
with user-specified weights $\pi = (\pi_1, \ldots, \pi_N) \in \mathring{\Delta}^{N-1}$. 

\subsubsection{Transport plan approximation}
In the empirical setting, exact Monge maps from the barycenter $\bar{\mu}$ to each input measure $\mu_n$ may not typically exist due to discreteness and potential mismatch in support cardinalities. To circumvent this, we replace transport maps with optimal transport plans, followed by their approximation via barycentric projection \citep{peyre_2019_ComputationalOptimalTransport}.

Let the current barycenter estimate be given by
\begin{equation*}
    \bar{\mu} = \sum_{i=1}^m v_i \delta_{z_i},
\end{equation*}
for a probability vector $v = (v_1,\ldots,  v_m)$ and free-support points $\lbrace z_i\rbrace_{i=1}^m \subset \bbR^d$. For each measure $\mu_n = \sum_{j=1}^{m_n} w_{n,j} \delta_{x_{n,j}}$, we solve the discrete optimal transport problem 
\begin{equation*}
    W_2^2 (\bar{\mu}, \mu_n) = \underset{\Gamma^{(n)}\in \Pi(v, w_n)}{\min}~ \sum_{i=1}^m \sum_{j=1}^{m_n} \Gamma^{(n)}_{i,j}  \|z_i - x_{n,j}\|^2,
\end{equation*}
where $\Gamma^{(n)} \in \bbR^{m\times m_n}$ is a nonnegative matrix with prescribed row and column sums $v$ and $w_n$ respectively. The attained minimizer $\hat{\Gamma}^{(n)}$ represents an optimal transport plan between $\bar{\mu}$ and $\mu_n$ according to the Kantorovich formulation.

Unfortunately, this does not  yield a deterministic map from each $z_i$ to a point in $\mu_n$. The barycentric projection is a standard surrogate that approximates the map by averaging over the support of $\mu_n$. For each barycenter point $z_i^{(k)}$, the projection under $\hat{\Gamma}_n$ is defined as 
\begin{equation*}
    T_n(z_i) = \frac{1}{v_i} \sum_{j=1}^{m_n} \hat{\Gamma}^{(n)}_{i,j} x_{n,j}.
\end{equation*}
This quantity represents the conditional expectation of the target locations given the transport mass departing from $z_i$, and hence serves as a continuous proxy for the potentially non-existent Monge map from $\bar{\mu}$ to $\mu_n$.  We then define the approximate logarithmic map from $\bar{\mu}$ to $\mu_n$ as the displacement vector
\begin{equation*}
    \widetilde{\log}_{\bar{\mu}}^{(n)}(z_i) := T_n(z_i) - z_i,
\end{equation*}
which captures the averaged direction and magnitude by which mass should be moved from $z_i^{(k)}$ to optimally reach $\mu_n$ under the plan $\hat{\Gamma}^{(n)}$. We use these projected displacements to approximate the Riemannian gradient of the barycenter functional
\begin{equation*}
    \widetilde{\text{grad}}_{\bar{\mu}} := -2 \sum_{n=1}^N \pi_n \widetilde{\log}_{\bar{\mu}}^{(n)}.
\end{equation*}
Each step in our optimization will update the barycenter support $\lbrace z_i\rbrace$ using this approximate gradient, consistent with the Riemannian structure.

\subsubsection{Riemannian gradient update via barycentric projection}
Having constructed an approximate Riemannian gradient, we now specify the full iterative update rule to compute the free-support Wasserstein barycenter. Let $\bar{\mu}^{(k)} = \sum_{i=1}^m v_i \delta_{z_i^{(k)}}$ be the current estimate of the barycenter at iteration $k$ with fixed weights $v = (v_1, \ldots, v_m)$. For each $\mu_n$, denote the optimal transport plan between $\bar{\mu}^{(k)}$ and $\mu_n$ as $\hat{\Gamma}^{(n,k)}$. Under this notation, each support point $z_i^{(k)}$ is associated with the barycentric projection $T_n^{(k)}$ and the approximate displacement vector $\widetilde{\log}_{\bar{\mu}^{(k)}}^{(n)}$ as 
\begin{equation*}
  T_n^{(k)}(z_i^{(k)}) := \frac{1}{v_i} \sum_{j=1}^{m_n} \hat{\Gamma}_{i,j}^{(n,k)} x_{n,j}\quad\text{and}\quad \widetilde{\log}_{\bar{\mu}^{(k)}}^{(n)}(z_i^{(k)}):= T_n^{(k)}(z_i^{(k)}) - z_i^{(k)}.
\end{equation*}
To update the support locations, we move in the direction of the averaged displacement across all measures, scaled by the gradient step size $\eta_k >0$:
\begin{equation}
    \begin{aligned}
    z_i^{(k+1)} &= z_i^{(k)} + 2\eta_k \sum_{n=1}^N \pi_n (T_n^{(k)} (z_i^{(k)}) - z_i^{(k)}) \\ 
    &= (1-2\eta_k)z_i^{(k)} + 2\eta_k \sum_{n=1}^N \pi_n T_n^{(k)} (z_i^{(k)}) \\ 
    &= (1-2\eta_k)z_i^{(k)} + 2\eta_k \sum_{n=1}^N \pi_n \left(\frac{1}{v_i} \sum_{j=1}^{m_n} \hat{\Gamma}_{i,j}^{(n,k)} x_{n,j}\right),
    \end{aligned}
\end{equation}
for all $i\in [m]$. This expression can be interpreted as a first-order step in the Wasserstein space using the approximate gradient direction obtained by barycentric projections. The scheme preserves the convex combination structure and maintains compatibility with the manifold geometry of $\calP_2(\bbR^d)$.

As proposed by \cite{zemel_2019_FrechetMeansProcrustes}, an effective choice of the step size is $\eta_k = 1/2$, which simplifies the update to 
\begin{equation*}
    z_i^{(k+1)} = \sum_{n=1}^N \pi_n T_n^{(k)}(z_i^{(k)}),
\end{equation*}
which implies that the new support point is the weighted average of its barycentric projections under the optimal plans. This iteration is repeated until convergence, yielding an estimate of the Wasserstein barycenter with freely updated support locations $\lbrace z_i\rbrace$.

\subsubsection{Remarks on convergence and implementation}

The algorithm developed in this section constitutes a Riemannian gradient descent scheme adapted to the computation of Wasserstein barycenters with free support. The update rule is consistent with the Otto calculus framework, with gradient steps taken via approximate logarithmic maps constructed from barycentric projections. In this subsection, we discuss theoretical convergence properties and practical considerations.

The convergence of the algorithm is inherited from the general structure of Riemannian gradient descent. Specifically, under mild conditions, such as compactness of the support and Lipschitz continuity of the cost function, each iteration yields a monotone decrease in the objective
\begin{equation*}
    \calF(\bar{\mu}^{(k+1)})\leq \calF(\bar{\mu}^{(k)}).
\end{equation*}
This holds particularly when the step size $\eta_k \in (0,1/2]$ is chosen appropriately. For the smooth map setting, $\eta_k = 1/2$ maximizes the one-step decrease \citep{zemel_2019_FrechetMeansProcrustes}. In our free-support scheme with barycentric projections, $\eta_k \in (0, 1/2]$ yields monotone descent under the Lipschitz assumptions below with $1/2$ working well empirically.

Although global convergence to a unique barycenter is not guaranteed due to nonconvexity of the objective in the free-support setting, empirical evidence suggests that the algorithm converges reliably when initialized with reasonable support points. For measures with well-separated modes or strong symmetry, local minima can be informative prototypes in unsupervised settings.

We offer a few remarks on implementation. The optimal transport plan $\Gamma^{(n,k)}$ must be calculated in each iteration between the current barycenter $\bar{\mu}^{(k)}$ and each input measure $\mu_n$. This accounts for solving multiple standard linear programs over the product of simplices independently, which can be done in parallel. For each subproblem, there exist efficient solvers such as network simplex \citep{bazaraa_2010_LinearProgrammingNetwork}, or even approximation of such if entropic regularization is allowed \citep{cuturi_2013_SinkhornDistancesLightspeed}. The quality of initialization is significant for such a nonconvex optimization. We conjecture that the barycenter of multiple probability measures should reflect the overall shape of the union of these measures. In line with this reasoning, we propose to use the $k$-means centers of all pooled support points from $\lbrace \mu_n \rbrace$. For termination of the updates, we employ a convergence criterion based on relative change in the objective value, i.e., 
\begin{equation*}
    \frac{|\calF(\bar{\mu}^{(k+1)})-\calF(\bar{\mu}^{(k)})|}{\calF(\bar{\mu}^{(k)})} < \varepsilon,
\end{equation*}
for some small tolerance $\varepsilon > 0$.

\section{Theory}\label{sec:chp4-theory}

\subsection{Approximation guarantees of barycentric projection}

As noted earlier, the Monge map from a source to a target measure may not exist for discrete measures, especially when the supports differ in cardinality. As a surrogate, we use the barycentric projection associated with the optimal transport plan, which defines a map from the source support to convex combinations of target support points. We formalize the approximation behavior of the barycentric projection in the limit where the empirical source measures converge to a continuous one. 

Let $\nu$ be a fixed absolutely continuous probability measure on $\bbR^d$ and $\bar{\mu}_m  = \sum_{i=1}^m v_i \delta_{z_i}$ be a sequence of empirical measures that weakly converge to an absolutely continuous measure $\bar{\mu}$. Denote $\Gamma_m$ for the optimal transport plan from $\bar{\mu}_m$ to $\nu$. The barycentric projection associated with $\Gamma_m$ is defined for each $z_i$ as 
\begin{equation*}
    T_m(z_i):= \frac{1}{v_i} \int_{\bbR^d} x d\Gamma_m(x\mid z_i),
\end{equation*}
where $\Gamma_m(x\mid z_i)$ is the conditional distribution over $\nu$'s support given the departure point $z_i$. The approximate logarithmic map is then given by 
\begin{equation*}
    \widetilde{\log}_{\bar{\mu}_m}^{(\nu)}(z_i):= T_m(z_i)-z_i.
\end{equation*}
We aim to show that this quantity converges in $L^2$ sense to the true logarithmic map $\log_{\bar{\mu}}(\nu)$ defined by the Monge map $T$ from $\bar{\mu}$ to $\nu$ as $\log_{\bar{\mu}}(\nu)(z) = T(z)-z$, assuming the existence of such map.

\begin{lemma}\label{theory:consistency_barycentric_projection}
    Let $\bar{\mu}_m = \sum_{i=1}^m v_i \delta_{z_i}$ be a weakly convergent sequence of measures to $\bar{\mu} \in \calP_2(\bbR^d)$, which is absolutely continuous with compact support. Let $\nu \in \calP_2(\bbR^d)$ be absolutely continuous with compact support. Denote by $T_m(z_i)$ the barycentric projection induced by the optimal transport plan 
    $\Gamma_m \in \Pi(\bar{\mu}_m, \nu)$, and $T$ the Brenier map from $\bar{\mu}$ to $\nu$. Then, 
    \begin{equation*}
        \lim_{m\to\infty} \int_{\bbR^d} \left\|\widetilde{\log}_{\bar{\mu}_m}^{(\nu)}(z_i) - \log_{\bar{\mu}}(\nu)(z_i)\right\|^2 d\bar{\mu}_m(z_i) = 0.
    \end{equation*}
\end{lemma}
\begin{proof}
Let $\gamma = (\text{Id},T)\#\bar{\mu} \in \Pi(\bar{\mu}, \nu)$ be the Monge coupling induced by the optimal map $T$. Since $\bar{\mu}_m \rightharpoonup \bar{\mu}$ and both $\bar{\mu}, \nu$ are absolutely continuous with compact support, standard stability results in optimal transport ensure that $\Gamma_m \rightharpoonup \gamma$ weakly \citep{villani_2003_TopicsOptimalTransportation}.

Let us fix a test function $\varphi \in C_b(\bbR^d\times \bbR^d)$ in the set of continuous and uniformly bounded functions. The weak convergence $\Gamma_m \rightharpoonup \gamma$ implies that 
\begin{equation*}
    \int \varphi(z,y)d\Gamma_m(z,y) \to \int \varphi(z,y) d\gamma(z,y).
\end{equation*}
Set $\varphi(z,y) = \|z-y\|^2$, which is continuous and bounded on compact sets. Then, 
\begin{equation*}
    \int \|T_m(z_i)-z_i\|^2 d\bar{\mu}_m(z_i) = \int \|y-z\|^2 d\Gamma_m(z,y) \to \int \|T(z) -z\|^2 d\bar{\mu}(z).
\end{equation*}
Now define the error
\begin{equation*}
    E_m := \int \left\|\widetilde{\log}_{\bar{\mu}_m}^{(\nu)}(z_i) - \log_{\bar{\mu}}(\nu)(z_i) \right\|^2 d\bar{\mu}_m (z_i),
\end{equation*}
which equals 
\begin{equation*}
    E_m = \int \|T_m(z_i) - T(z_i)\|^2 d\bar{\mu}_m(z_i).
\end{equation*}
In order to bound this, we invoke the convergence of barycentric projections in $L^2$ 
\begin{equation*}
    \int \|T_m(z_i) - T(z_i)\|^2 d\bar{\mu}_m(z_i) \to 0,
\end{equation*}
under the compact support and uniform integrability of the second moments assumption \citep{peyre_2019_ComputationalOptimalTransport}. By recognizing the equivalent forms of the error $E_m$, this completes the proof. 
\end{proof}

Lemma \ref{theory:consistency_barycentric_projection} provides a theoretical justification for using barycentric projections as an approximation to the Riemannian gradient in Wasserstein space. It shows that, as the empirical input measures become dense enough, the approximate gradient computed from the transport plans converges to the true gradient direction in $L^2$ sense. This result directly connects to the known results on the convergence of the barycentric projections of optimal transport plans to the Monge map under suitable conditions \citep{peyre_2019_ComputationalOptimalTransport}. In our setting, it means that updating the barycenter using these projections remains consistent with the true gradient flow, making the discrete algorithm a reliable approximation of the continuous Wasserstein gradient descent. This convergence provides an effective approximation of the continuous gradient flow, ensuring that the algorithm maintains its reliability in large-scale computations and empirical settings. 

\subsection{Convergence of the free-support barycenter problem}

We now turn to the convergence properties of the iterative scheme described in Section \ref{sec:chp3-methods}. Recall that each iteration consists of a Riemannian gradient step with respect to the support points. We provide a convergence result for the objective functional under natural regularity conditions and offer justification through both geometric insights and first-order analysis.

Let us denote the barycenter functional by 
\begin{equation*}
    \calF(\bar{\mu}) = \sum_{n=1}^N \pi_n W_2^2(\bar{\mu}, \mu_n),
\end{equation*}
where each $\mu_n = \sum_{j=1}^{m_n} w_{n,j} \delta_{x_{n,j}}$ is an empirical measure with finite second moment and compact support in $\bbR^d$. Suppose that the initial barycenter estimate $\bar{\mu}^{(0)} = \sum_{i=1}^m v_i^{(0)} \delta_{z_i^{(0)}}$ has its support contained in a compact convex subset $K \subset\bbR^d$. The algorithm generates a sequence $\bar{\mu}^{(k)}$ by updating the support points via barycentric projections.

We denote the approximate Riemannian gradient of $\calF$ with respect to the support location $z_i$ by 
\begin{equation*}
    \tilde{\nabla}_{z_i}\calF := -2 \sum_{n=1}^N \pi_n (T_n(z_i)-z_i),
\end{equation*}
where $T_n(z_i)$ is the barycentric projection of $z_i$ under the optimal plan $\Gamma^{(n)}$ from the current iterate $\barmu^{(k)}$ to $\mu_n$. The update rule for each $z_i$ is given by 
\begin{equation*}
    z_i^{(k+1)} = z_i^{(k)} + 2\eta_k \sum_{n=1}^N \pi_n (T_n^{(k)} (z_i^{(k)}) - z_i^{(k)}),
\end{equation*}
for the step size $\eta_k \in (0, 1/2]$. We now state the main convergence result.

\begin{theorem}\label{theory:monotonic_descent_and_convergence}
    Assume the support points $\{z_i^{(k)}\}_{i=1}^m$ remain in a compact convex set $K \subset \mathbb{R}^d$ and that for each $n$, the optimal plan $\Gamma^{(n,k)}$ between $\mu^{(k)}$ and $\mu_n$ is locally unique along the iterate path and the active LP basis remains constant except on a finite union of switching hyperplanes so that $Z \mapsto T_n^{(k)}(z_i)$ is piecewise affine, thus Lipschitz on compact subsets that avoid basis switches. Then the sequence of objective values $\{\mathcal{F}(\bar{\mu}^{(k)})\}$ is non-increasing and converges. Moreover, any limit point of the sequence $\bar{\mu}^{(k)}$ is a stationary point of $\mathcal{F}(\bar{\mu})$.
\end{theorem}

\begin{proof}
    We start by arguing that the approximate gradient $\widetilde{\nabla}_{z_i}\mathcal{F}$ is Lipschitz continuous in the support points. Each $T_n(z_i)$ is defined as the conditional expectation under the transport plan $\Gamma^{(n)}$, i.e.,
    \begin{equation*}
        T_n(z_i) = \frac{1}{v_i} \sum_{j=1}^{m_n} \Gamma_{i,j}^{(n)} x_{n,j}.
    \end{equation*}
    As the entries of $\Gamma^{(n)}$ are obtained by solving a linear program over the product of simplices, and the cost matrix varies smoothly with $z_i$, it follows that $\Gamma^{(n)}$ depends continuously and piecewise smoothly on $\{z_i\}$. Therefore, each $T_n(z_i)$ is a piecewise smooth function of $z_i$, and is Lipschitz continuous over compact domains. This implies that the approximate gradient $\widetilde{\nabla}_{z_i} \mathcal{F}$, which is a finite weighted sum of such projections, is also piecewise Lipschitz, i.e., on any compact subset $K'\subset K$ that avoids basis switches the Lipschitz constant is finite. The descent argument applies on each of such regions. When a switch occurs, the objective is continuous and the inequality resumes on the next regions. Moreover, since the optimal transport plans $\Gamma^{(n,k)}$ are assumed to be \emph{locally stable and unique}, the corresponding barycentric projections $T_n^{(k)}(z_i^{(k)})$ depend Lipschitz-continuously on the support points in a neighborhood around the current iterate. This ensures that the gradient of the objective function $\mathcal{F}(\bar{\mu}^{(k)})$ is Lipschitz continuous with respect to the support points.

    Let $Z^{(k)} = (z_1^{(k)}, \ldots, z_m^{(k)})$ denote the vector of barycenter support points at iteration $k$, and let $Z^{(k+1)} = Z^{(k)} - \eta_k \widetilde{\nabla}\mathcal{F}(Z^{(k)})$ be the next iterate. The first-order Taylor expansion of $\mathcal{F}$ at $Z^{(k)}$ gives
    \begin{equation*}
        \mathcal{F}(Z^{(k+1)}) = \mathcal{F}(Z^{(k)}) - \eta_k \|\widetilde{\nabla} \mathcal{F}(Z^{(k)})\|^2 + \mathcal{O}(\eta_k^2),
    \end{equation*}
    where the second-order term is bounded due to the Lipschitz continuity of $\widetilde{\nabla} \mathcal{F}$ and compactness of the domain. Therefore, for sufficiently small $\eta_k \in (0, \eta_{\max}]$, we have monotonic descent, i.e., 
    \[
    \mathcal{F}(Z^{(k+1)}) \leq \mathcal{F}(Z^{(k)}),
    \]
    and the sequence converges as $\mathcal{F}$ is bounded below by zero.

    Additionally, if the norm $\|\widetilde{\nabla} \mathcal{F}(Z^{(k)})\|$ does not vanish, the objective strictly decreases per the descent inequality. Hence, the only possible accumulation points of $\{Z^{(k)}\}$ must satisfy $\widetilde{\nabla} \mathcal{F} = 0$, which characterizes stationary points under the barycentric scheme.
\end{proof}

We note that Theorem \ref{theory:monotonic_descent_and_convergence} relies on two ingredients: compactness of the domain $K$ and the Lipschitz continuity of the barycentric projections. The assumption on compactness is natural in practical scenarios, where support points are initialized in a bounded region and updated via convex combinations, thus remaining in $K$. The Lipschitz continuity of the barycentric projections is a direct consequence of the smooth dependence of the optimal transport plan $\Gamma^{(n)}$ on the cost matrix, which itself depends quadratically on the support locations. While this version proves convergence qualitatively, a sharper quantitative descent rate can be obtained given the characterization of the Lipschitz constant.

\begin{corollary}\label{theory:monotonic_descent_and_rate_of_convergence}
    Under the assumptions of Theorem~\ref{theory:monotonic_descent_and_convergence}, suppose in addition that $\widetilde{\nabla} \calF$ is $L$-Lipschitz on a compact region that contains the iterates. Then, for any fixed step size $\eta \in (0, 2/L)$, the iterates satisfy the descent inequality
    \[
        \mathcal{F}(\bar{\mu}^{(k+1)}) \leq \mathcal{F}(\bar{\mu}^{(k)}) - \left( \eta - \frac{L\eta^2}{2} \right) \|\widetilde{\nabla} \mathcal{F}(\bar{\mu}^{(k)})\|^2.
    \]
    In particular, setting $\eta = 1/L$ gives the optimal bound
    \[
        \mathcal{F}(\bar{\mu}^{(k+1)}) \leq \mathcal{F}(\bar{\mu}^{(k)}) - \frac{1}{2L} \|\widetilde{\nabla} \mathcal{F}(\bar{\mu}^{(k)})\|^2.
    \]
\end{corollary}
\begin{proof}
    Under the assumptions of Theorem~\ref{theory:monotonic_descent_and_convergence}, we know that the gradient $\widetilde{\nabla} \mathcal{F}(\bar{\mu}^{(k)})$ is Lipschitz continuous, as the barycentric projections \(T_n^{(k)}(z_i^{(k)})\) are locally Lipschitz in \(z_i^{(k)}\). This allows us to apply standard gradient descent analysis.

    Since $\mathcal{F}(\bar{\mu})$ is continuously differentiable and $\widetilde{\nabla} \mathcal{F}(\bar{\mu})$ is Lipschitz continuous, the sequence $\mathcal{F}(\bar{\mu}^{(k)})$ converges to a stationary point at a rate governed by the step size $\eta$. Specifically, for any fixed step size $\eta \in (0, 2/L)$, we have the descent inequality
    \[
        \mathcal{F}(\bar{\mu}^{(k+1)}) \leq \mathcal{F}(\bar{\mu}^{(k)}) - \left( \eta - \frac{L\eta^2}{2} \right) \|\widetilde{\nabla} \mathcal{F}(\bar{\mu}^{(k)})\|^2.
    \]
    In particular, setting $\eta = 1/L$ gives the optimal bound
    \[
        \mathcal{F}(\bar{\mu}^{(k+1)}) \leq \mathcal{F}(\bar{\mu}^{(k)}) - \frac{1}{2L} \|\widetilde{\nabla} \mathcal{F}(\bar{\mu}^{(k)})\|^2.
    \]
\end{proof}

This result enhances the earlier convergence guarantee by providing a quantitative estimate of descent at each iteration in terms of the norm of the approximate gradient. The Lipschitz assumption is standard in optimization and is satisfied here due to the compactness of the support set $K$ and the smooth dependence of the barycentric projections on the support points. The bound shows that even though the gradient is only approximate via a barycentric projection, the algorithm behaves like a classical gradient descent under smoothness.

The optimal step size $\eta = 1/L$ yields the sharpest decrease per step. This is consistent with the general theory of smooth optimization, as discussed in \cite{nesterov_2018_LecturesConvexOptimization}. Furthermore, the coefficient $(\eta - L\eta^2/2)$ governs the convergence speed, and the iterations converge to stationary points in the limit of vanishing gradients, which provide a more rigorous foundation for the monotonic behavior of the barycenter functional in our discrete free-support setting.

\subsection{Stability of the barycenter with respect to input measures}

The Wasserstein barycenter inherits a degree of stability from the continuity properties of the Wasserstein distance. We establish that small perturbations in the input measures lead to small perturbations in the resulting free-support barycenter, both in terms of the objective value and the location of support points. This is particularly relevant in empirical settings where the observed measures $\lbrace \mu_n\rbrace_{n=1}^N$ are approximations of the population distributions $\lbrace \nu_n\rbrace_{n=1}^N$. We show that if each $\mu_n$ is close to $\nu_n$ in the $W_2$ sense, so are the value of the barycenter functional and its output location.

Let $\lbrace \nu_n\rbrace_{n=1}^N$ denote the original ground-truth measures and $\lbrace \mu_n\rbrace_{n=1}^N$ the perturbed versions. Fix the weights $\pi=(\pi_1,\ldots,\pi_N)\in\mathring{\Delta}^{N-1}$ and define the corresponding barycenter functionals
\begin{equation*}
    \calF_\nu(\bar{\mu}):= \sum_{n=1}^N \pi_n W_2^2(\bar{\mu}, \nu_n),\quad \calF_\mu(\bar{\mu}) := \sum_{n=1}^N \pi_n W_2^2 (\bar{\mu}, \mu_n).
\end{equation*}
Let $\bar{\mu}^*$ and $\bar{\mu}_m$ denote a minimizer of $\calF_\nu$, and  the output of the barycentric-projection gradient descent applied to $\lbrace \mu_n\rbrace$ under fixed support size $m$. Our goal is to assess the difference between $\calF_\nu(\bar{\mu}_m)$ and $\calF_\nu(\bar{\mu}^*)$ for quantifying the robustness of the barycenter algorithm.

\begin{proposition}\label{theory:stability_barycenter}
    Assume that all measures $\mu_1, \ldots, \mu_N$ and their barycenter $\bar{\mu}_m$ are supported within a common compact ball $B(0, R)$ for some $R > 0$. Let $\delta = \max_{n \in [N]} W_2(\mu_n, \nu_n)$ denote the uniform upper bound on the perturbations in the Wasserstein distance. Then the difference between the perturbed and true barycenter functionals satisfies
    \[
        |\mathcal{F}_\mu(\bar{\mu}_m) - \mathcal{F}_\nu(\bar{\mu}_m)| \leq 4R\delta,
    \]
    for any candidate measure $\bar{\mu} \subset B(0, R)$.
\end{proposition}
\begin{proof}
    Start by expanding the absolute difference as follows:
    \begin{equation*}
        |\mathcal{F}_\mu(\bar{\mu}_m) - \mathcal{F}_\nu(\bar{\mu}_m)| = \left|\sum_{n=1}^N \pi_n \left(W_2^2(\bar{\mu}_m, \mu_n) - W_2^2(\bar{\mu}_m, \nu_n)\right)\right| \leq \sum_{n=1}^N \pi_n |W_2^2(\bar{\mu}_m, \mu_n) - W_2^2(\bar{\mu}_m, \nu_n)|.
    \end{equation*}
    For each term, we have:
    \begin{align*}
        |W_2^2(\bar{\mu}_m, \mu_n) - W_2^2(\bar{\mu}_m, \nu_n)| &= |W_2(\bar{\mu}_m, \mu_n) + W_2(\bar{\mu}_m, \nu_n)| \cdot |W_2(\bar{\mu}_m, \mu_n) - W_2(\bar{\mu}_m, \nu_n)| \\
        &\leq |W_2(\bar{\mu}_m, \mu_n) + W_2(\bar{\mu}_m, \nu_n)| \cdot W_2(\mu_n, \nu_n) \\
        &\leq \delta \cdot \left(W_2(\bar{\mu}_m, \mu_n) + W_2(\bar{\mu}_m, \nu_n)\right),
    \end{align*}
    where the last inequality comes from the uniform upper bound \(\delta = \max_{n \in [N]} W_2(\mu_n, \nu_n)\) in the statement.

    Since all measures are supported on the compact ball \(B(0, R)\), we have the uniform bound:
    \[
    W_2(\bar{\mu}_m, \mu_n) \leq \text{diam}(B(0, R)) = 2R, \quad W_2(\bar{\mu}_m, \nu_n) \leq 2R.
    \]
    Hence, for each \(n\), we have:
    \[
    |W_2^2(\bar{\mu}_m, \mu_n) - W_2^2(\bar{\mu}_m, \nu_n)| \leq 4R\delta,
    \]
    which leads to the final bound:
    \[
    |\mathcal{F}_\mu(\bar{\mu}_m) - \mathcal{F}_\nu(\bar{\mu}_m)| \leq \sum_{n=1}^N \pi_n \cdot 4R\delta = 4R\delta,
    \]
    since \(\sum_{n=1}^N \pi_n = 1\).    Therefore, the difference between the perturbed and true barycenter functionals is bounded by \(4R\delta\), which completes the proof.
\end{proof}

This result demonstrates the quantitative robustness of the proposed algorithm to small perturbations in the input measures. Specifically, if each perturbed input measure remains within a 2-Wasserstein distance $\delta$ of its ground-truth counterpart and all measures are supported within a compact set of fixed radius, the objective value evaluated at the output barycenter remains within a specified tolerance from the ideal cost. This stability guarantee holds independently of the dimension or the barycenter's support cardinality. Moreover, it provides a principled bound on the propagation of the data noise or approximation error into the final barycentric estimate. We note that the compact support assumption  in Proposition~\ref{theory:stability_barycenter} can be relaxed. For instance, one can bound the difference in terms of second moments \citep{villani_2003_TopicsOptimalTransportation}. We conjecture that this gives the same type of stability guarantee without requiring bounded support.

\subsection{Consistency of barycentric support points under increasing resolution}

The algorithm requires the number of support points $m$ to be set a priori. This calls for theoretical support that if the algorithm is run with increasingly fine discretizations of the barycenter, i.e., larger support sizes, then the resulting measures converge to a minimizer of the population barycenter functional in the 2-Wasserstein sense.

Let $\mu_1, \ldots, \mu_N \in \calP_2(\bbR^d)$ be fixed input measures with weights $\pi = (\pi_1, \ldots, \pi_N)\in\mathring{\Delta}^{N-1}$. The Wasserstein barycenter $\bar{\mu}^*$ is the unique minimizer of the functional 
\begin{equation*}
    \calF(\mu) = \sum_{n=1}^N \pi_n W_2^2(\mu,\mu_n),\quad \mu\in \calP_2(\bbR^d).
\end{equation*}
Suppose the barycenter is approximated by restricting it to be a discrete measure supported on at most $m$ atoms, the estimate of which is denoted as $\bar{\mu}_m$. Our goal is to establish that $\bar{\mu}_m$ converges to the true barycenter as $m\to\infty$.

\begin{theorem}\label{theory:consistency_resolution}
    Assume \(\mu_n\) is compactly supported in a common bounded convex subset \(K \subset \mathbb{R}^d\) for all \(n \in [N]\), and there exists at least one index \(n'\) such that \(\mu_{n'}\) is absolutely continuous. Then, any sequence of discrete barycenters \(( \bar{\mu}_m)\) satisfying the minimizer criterion converges to the true barycenter \(\bar{\mu}^*\) in 2-Wasserstein distance, i.e., 
    \[
    W_2(\bar{\mu}_m, \bar{\mu}^*) \to 0 \quad \text{as} \quad m \to \infty.
    \]
    The same holds when \(\bar{\mu}_m\) is an approximate minimizer obtained via the barycentric-projection gradient algorithm, provided that the gradient norm tends to zero and the support remains in a compact subset of \(K\). Furthermore, the convergence is uniform with respect to the support size \(m\), as the objective value stabilizes and converges to the true barycenter.
\end{theorem}

\begin{proof}
    Since all \(\mu_n\)'s are supported on \(K\) and the squared Wasserstein cost is minimized over compactly supported candidates, each \(\bar{\mu}_m\) is also supported within \(K\). The sequence \(( \bar{\mu}_m)\) is therefore tight and uniformly integrable with uniformly bounded second moments. By Prokhorov's theorem, one can extract a subsequence \((\bar{\mu}_{m_k})\) that converges in \(W_2\) to a limit \(\bar{\mu}_\infty \in \mathcal{P}_2(K)\). 

    Lower semicontinuity of the functional \(\mathcal{F}\) under \(W_2\) convergence implies 
    \[
    \liminf_{k \to \infty} \mathcal{F}(\bar{\mu}_{m_k}) \geq \mathcal{F}(\bar{\mu}_\infty).
    \]
    Since each \(\bar{\mu}_{m_k}\) is a minimizer over its respective problem setup, we also have 
    \[
    \mathcal{F}(\bar{\mu}_{m_k}) \leq \mathcal{F}(\bar{\mu}^*).
    \]
    Therefore, 
    \[
    \mathcal{F}(\bar{\mu}_\infty) \leq \liminf_{k \to \infty} \mathcal{F}(\bar{\mu}_{m_k}) \leq \mathcal{F}(\bar{\mu}^*),
    \]
    and hence \(\bar{\mu}_\infty\) is also a global minimizer of \(\mathcal{F}\). Under the assumption that at least one of \(\mu_n\)'s is absolutely continuous, the barycenter is unique, so that \(\bar{\mu}_\infty = \bar{\mu}^*\). Since this holds for any converging subsequence, the entire sequence \(\bar{\mu}_m\) converges to \(\bar{\mu}^*\) in \(W_2\).

    The same conclusion holds when \(\bar{\mu}_m\) is merely an approximate output from the barycentric-projection gradient algorithm, as long as the following two conditions that (1) the norm of the approximate gradient converges to zero and (2) the support points remain within a compact subset of \(K\).
    The proof then follows by similar compactness and stability arguments using lower semicontinuity and the stability of Wasserstein barycenters under perturbations.
    
    Additionally, the convergence is uniform with respect to the support size \(m\) as the support points and the objective value stabilize when the resolution increases. As the number of support points increases, the sequence of barycenter estimates \(\bar{\mu}_m\) approaches the true barycenter \(\bar{\mu}^*\), ensuring that the approximation improves with higher resolution.
\end{proof}

Theorem \ref{theory:consistency_resolution} provides theoretical ground for an important empirical phenomenon of improved barycenter estimates with increasing resolution, expanding representational capacity of the discrete barycenters. This is especially meaningful in high-dimensional or structured data settings where enforcing strong parametric forms on the barycenter may be either inappropriate or restrictive. 

In practical terms, this consistency guarantees that any observed discrepancy between the computed barycenter and the true one can be attributed to under-resolution rather than algorithmic failure, provided that the support remains within a compact region and the optimization is performed accurately. This supports a principled strategy in algorithm design: begin with a coarse approximation, then progressively refine support size $m$, each time using the previous output as a warm start. Theoretical convergence ensures that this continuation approach converges to a meaningful solution without the need for external regularization or heuristics.

\section{Experiments}\label{sec:chp5-experiment}

In this section we evaluate the proposed free-support barycenter algorithm on synthetic and real problems that stress complementary aspects of accuracy, sharpness, and scalability. We organize the experiments into four parts: (1) Gaussian benchmarks with an oracle barycenter to quantify error versus support size,  (2) Bayesian posterior aggregation via the Wasserstein posterior, where we compare against full-data MCMC and examine how discretization affects location and covariance recovery, (3) prototype learning and nearest-prototype classification on MNIST digits using particle barycenters,  and (4) large-scale clustering through a newly proposed pipeline based on the quantization idea in which free-support barycenters serve as cluster prototypes.

\subsection{Gaussian barycenters}

The first simulated example benchmarks the proposed free-support barycenter against a population oracle available in closed form with respect to averaging of Gaussian distributions in the discrete setting. We generate four bivariate Gaussian distributions $\mu^{(i)}= \calN(m^{(i)}, \Sigma^{(i)}),~i=1,\ldots,4$, where the parameters are randomly drawn as $m^{(i)} \sim \calN(
10\cdot ((-1)^{\lfloor (i-1)/2 \rfloor},  (-1)^{i-1})^\top, I_2)$ for the $2\times 2$ identity matrix $I_2$ and  $\Sigma^{(i)} \sim Wishart(\textrm{df}=4, I_2)$, which we denote as components.  Then, we compute the oracle 2-Wasserstein barycenter $\mu^* = \calN(m^*, \Sigma^*)$ of $\lbrace \mu^{(i)} \rbrace_{i=1}^4$  using the fixed-point iteration under the Bures-Wasserstein metric \citep{takatsu_2011_WassersteinGeometryGaussian, alvarez-esteban_2016_FixedpointApproachBarycenters}. For each distribution from four types of target Gaussian measures, we sample 100 observations and denote the empirical counterparts as $\hat{\mu}^{(i)}$'s. Given these realizations, we estimate a discrete free-support barycenter $\bar{\mu}_n$ consisting of $n$ support points for $n \in \lbrace 10, 20, \ldots, 200\rbrace$ with uniform weights constraint for simplicity to investigate the impact of varying resolution in discrete barycenter estimation. We repeat the aforementioned procedure 100 times with random seeds.

\begin{figure}[ht]
	\centering
	\includegraphics[width=.9\textwidth]{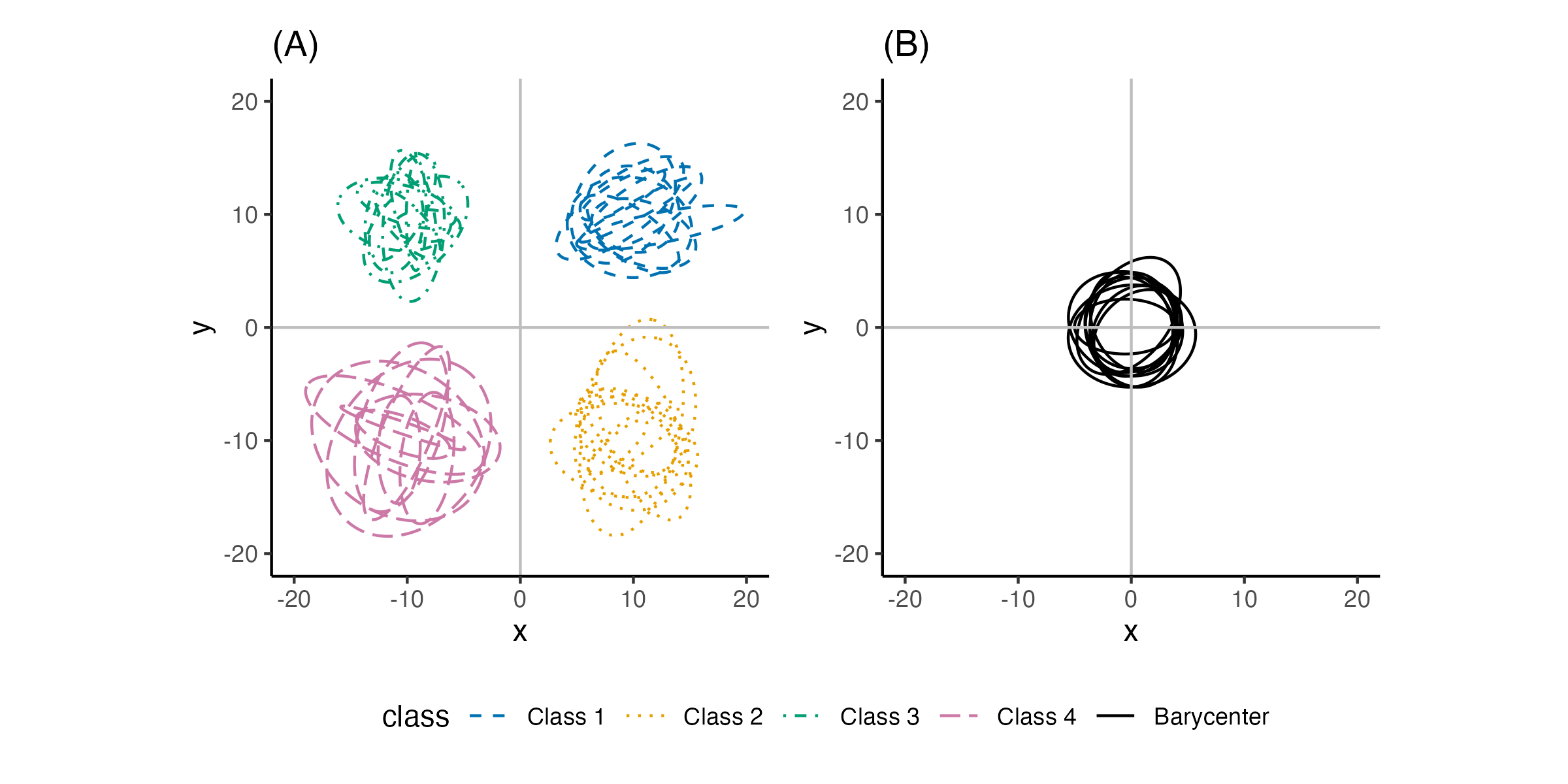}
	\caption{Two-dimensional Gaussian barycenter experiment. Shown are 95\% probability contours of confidence ellipses for (A) multiple realizations of four Gaussian components and (B) the corresponding oracle Gaussian barycenters.}
	\label{fig:simulation-gauss1}
\end{figure}

Figure \ref{fig:simulation-gauss1} shows random realizations of four component measures through confidence ellipses and their barycenters. These components are spatially dispersed in a symmetric manner in that their central location is expected to be near zero. Similarly, all covariances are drawn from the Wishart distribution with an identity scale matrix and the degrees of freedom of 4. The choice of Wishart parameters guarantees positive definiteness of a draw while ensuring certain degree of heterogeneity within the generated samples in terms of eccentricity and direction.  Visualization of oracle barycenters $\mu^*$ supports these ad hoc speculations that they are indeed centered around the origin and almost concentric. 

We evaluate the performance of free-support algorithm as follows. We first consider the semi-discrete setting to compute the Wasserstein distance between an oracle $\mu^*$ and an estimated barycenter $\bar{\mu}_n$ for varying $n$. The distance can be easily approximated by a standard Monte Carlo procedure where a random sample of size 100 is repeatedly drawn from $\mu^*$ and the distances between each sample and $\bar{\mu}_n$ are averaged. Next, we compare moments between the oracle and an empirical barycenter. We compute sample mean $\hat{m}_n$ and covariance $\hat{\Sigma}_n$  from  $\bar{\mu}_n$ and measure discrepancy between the MLEs and the oracle parameters $(m^*, \Sigma^*)$ by the standard Euclidean distance and the Bures-Wasserstein distance for each parameter, respectively. 

\begin{figure}[ht]
	\centering
	\includegraphics[width=.8\textwidth]{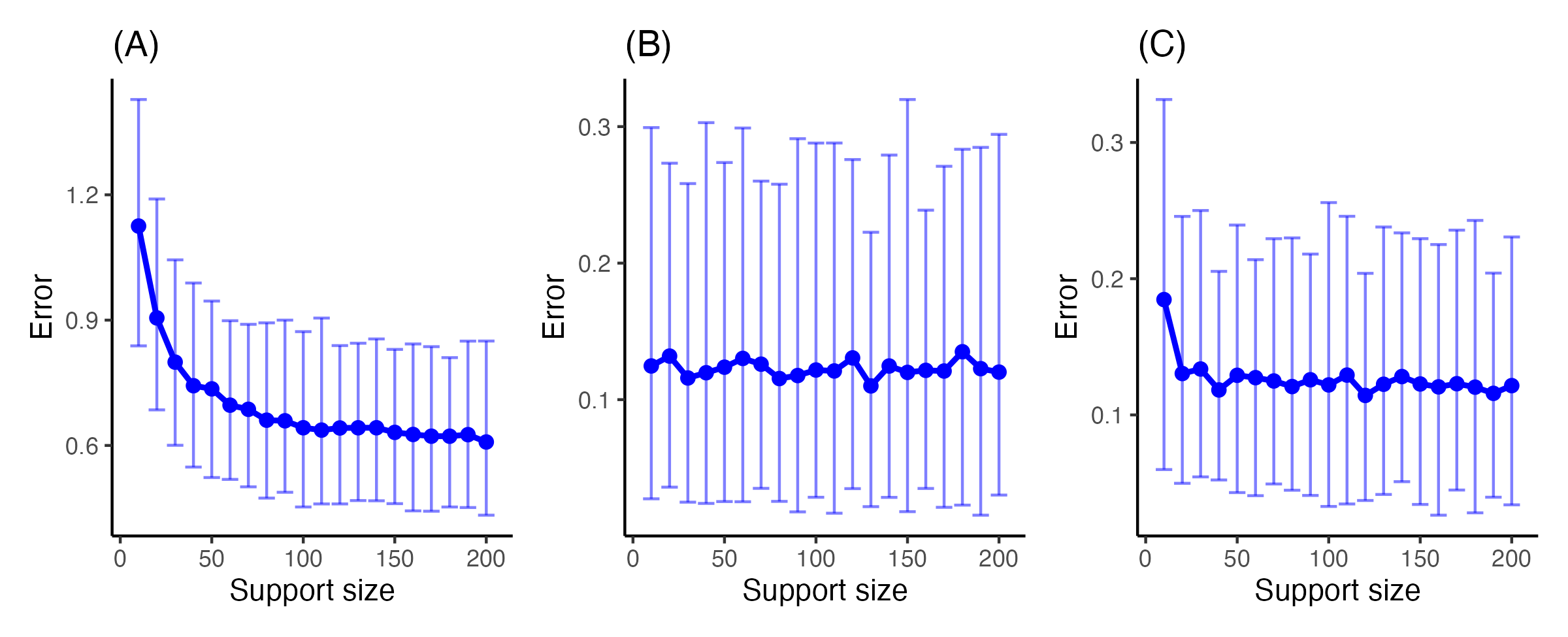}
	\caption{Free-support barycenter accuracy versus support size. 
		(A) Semi-discrete $W_2$ error between the oracle Gaussian barycenter $\mu^\star$ and the empirical free-support barycenter $\bar\mu_n$, estimated by Monte Carlo with 100 draws from $\mu^\star$ per repetition.
		(B) Mean error $\|\hat m_n - m^\star\|_2$. 
		(C) Covariance error $d_{\text{Bures}}(\hat\Sigma_n,\Sigma^\star)$. 
		Points show averages over 100 repetitions and vertical bars indicate 95\% Monte Carlo confidence intervals.}
	\label{fig:simulation-gauss2}
\end{figure} 

These performance metrics are shown in Figure \ref{fig:simulation-gauss2}, showing mean errors and 95\% Monte Carlo confidence intervals for each metric. It shows a clear decrease of the $W_2$ error as the support size grows, with most of the gain occurring for $n \leq 50$ and only modest improvement thereafter, which is consistent with a diminishing discretization bias and a residual floor due to finite data.  For the moments comparison, an interesting pattern is observed that the barycenter's mean is already well captured at small $n$. This is expected given the symmetry of the configuration and the averaging effect of the Bures-Wasserstein barycenter algorithm. Another observation is that there is a rapid early reduction of the covariance discrepancy, stabilizing once $n$ reaches a few dozen atoms. Beyond that, variability across repetitions, driven by the Wishart-drawn covariances and finite samples, dominates. Taken together, these curves suggest that moderate support sizes on the order of a few tens of atoms capture most of the improvement. It is worth mentioning that the remaining $W_2$ gap seems to be largely due to second-order component rather than location error. 

\subsection{Bayesian linear regression}

The second simulated example is Bayesian linear regression. In the literature of scalable Bayesian inference, the Wasserstein Posterior (WASP) is a divide-and-conquer strategy that partitions the dataset into multiple non-overlapping subsets, draws posterior samples on each subset with the likelihood appropriately rescaled, and combines the subset posteriors through their Wasserstein barycenter \citep{srivastava_2018_ScalableBayesBarycenter}. The resulting barycenter distribution serves as an approximation to the full-data posterior while alleviating the computational and memory burden associated with fitting a single MCMC on all data. 

We consider the following model. The data are generated from a two-dimensional linear regression,
\begin{equation*}
	y_i = x_i^\top \beta + \epsilon,\quad \epsilon\sim \mathcal{N}(0, \sigma^2),
\end{equation*}
with fixed parameters of $\beta = (1,1)^\top$ and $\sigma^2 = 1$ for $i = 1, 2, \ldots, n$. Covariates $x_i \in \bbR^2$ are drawn independently from a standard normal distribution and stacked row-wise into  $X \in \bbR^{n \times 2}$. We place a Gaussian prior 
\begin{equation*}
	\beta \sim \calN(m_0, V_0),\quad m_0 = (0,0)^\top, \, V_0 = 16 \cdot I_2,
\end{equation*}
which is relatively weakly informative. For this conjugate setting, the posterior distribution is available in closed form
\begin{equation*}
	\beta \mid (X,y) \sim \mathcal{N}(m_*, V_*),
\end{equation*}
with parameters given by 
\begin{equation*}
	V_* = \left(V_0^{-1} + \frac{1}{\sigma^2} X^\top X\right)^{-1},\quad m_* = V_*(V_0^{-1} m_0 + \frac{1}{\sigma^2} X^\top y).
\end{equation*}
We refer to this exact posterior as the oracle, which provides a ground-truth benchmark against which approximate methods can be evaluated.

In our experiment, we generate $n=10^4$ samples from this model. The oracle posterior is computed analytically, and a full-data MCMC posterior is also computed for baseline comparison. To approximate, the data are randomly split into $K$ subsets, $K \in \lbrace 2,3,\ldots, 20\rbrace$, each of nearly equal size. For a fixed $K$, the $k$-th subset posterior is fit via MCMC under a power likelihood with an exponent $\alpha = n/n_k$ to correct for sample size, where $n_k$ is the size of the $k$-th subset. The resulting subset posteriors are aggregated into a WASP approximation by computing empirical Wasserstein barycenters with support sizes varying from 10 to 200. Each configuration is repeated 100 times to quantify Monte Carlo variability. This design enables systematic evaluation of how the number of subsets, stochastic error, and barycenter resolution affect the fidelity of WASP approximations relative to the oracle posterior.

For posterior sampling, we use the No-U-Turn Sampler (NUTS) \citep{hoffman_2014_NoUTurnSamplerAdaptively} as implemented in the \textsf{Stan} probabilistic programming language \citep{standevelopmentteam_2025_StanReferenceManual}. We run 4 chains per model, each with 2{,}000 warmup iterations followed by 20,000 sampling iterations, thinned by 100 and thus retaining 200 posterior draws per chain. A single chain is chosen at random and its samples are used. Diagnostics $\widehat{R}$ and effective sample sizes (ESS) are monitored in that we require $\widehat{R}<1.01$ for all scalar parameters and both bulk- and tail-ESS $\ge 400$. Runs failing any criterion are rerun with a higher \texttt{adapt\_delta} parameter and, if needed, increased iterations. The full-data MCMC baseline is fit with the same settings and thresholds.

\begin{figure}[ht]
	\centering
	\includegraphics[width=.95\textwidth]{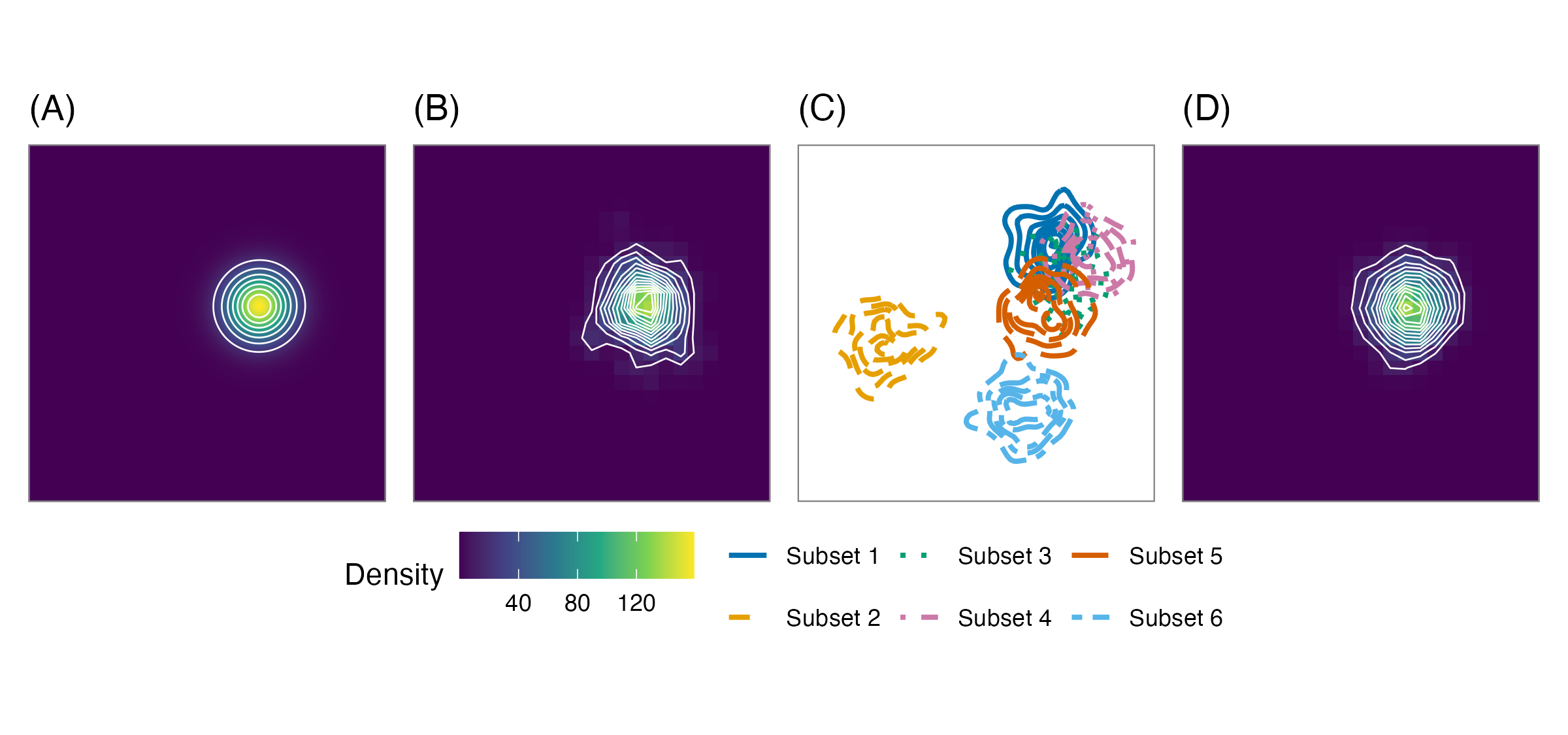}
	\caption{Kernel density estimates of posterior distributions. (A) oracle posterior in closed form, (B) full-data MCMC posterior, (C) subset posteriors with $K=6$, and (D) WASP barycenter of subset posteriors. All are plotted on the same scale for comparison.}
	\label{fig:simulation-normal1}
\end{figure} 

Figure~\ref{fig:simulation-normal1} displays a representative run with $K=6$. The full-data MCMC posterior closely matches the oracle, while the subset posteriors exhibit visible dispersion and shifts in location. Aggregating them through a Wasserstein barycenter successfully recovers the overall shape, yielding a smoother and more regular approximation that remains faithful to the oracle distribution.

\begin{figure}[ht]
	\centering
	\includegraphics[width=.95\textwidth]{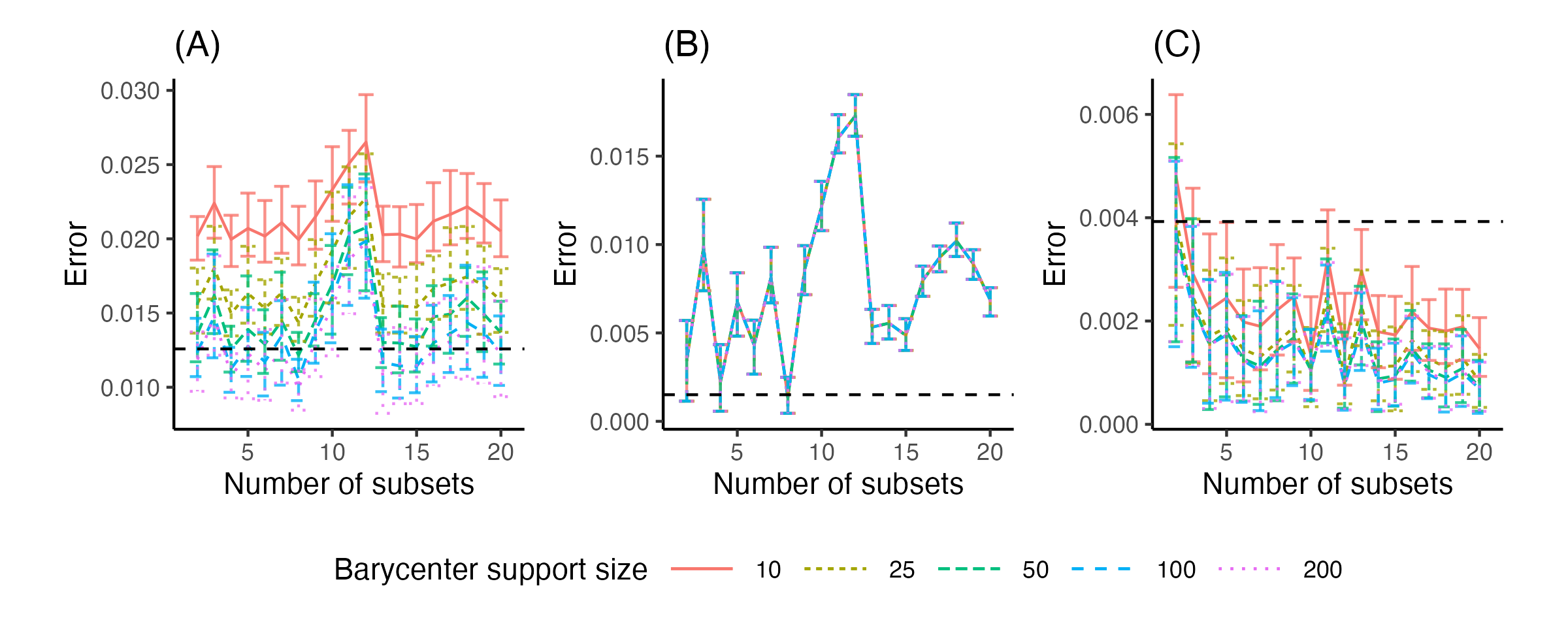}
	\caption{Accuracy of WASP across experimental settings. (A) semi-discrete Wasserstein error relative to the oracle, (B) error in posterior mean, and (C) error in posterior covariance. Lines correspond to barycenter support sizes ${10,25,50,100,200}$, with error bars indicating Monte Carlo variability across 100 replications. Black dashed lines denote the corresponding errors of the full-data MCMC posterior.}
	\label{fig:simulation-normal2}
\end{figure} 

Figure~\ref{fig:simulation-normal2} summarizes the results. Larger barycenter supports consistently reduce error, particularly for the covariance and the Wasserstein discrepancy. This pattern mirrors our earlier simulation, with location estimates remaining relatively stable and covariance driving most of the variation in accuracy. Notably, the full-data MCMC sample, represented as black dashed lines in the figure, achieves the lowest error for the posterior mean but performs comparatively poorly in covariance recovery, placing it near the worst end of the spectrum across all configurations. In contrast, WASP with sufficiently large support attains smaller covariance error while losing fidelity in mean error by small margins, leading to a lower overall semi-discrete discrepancy. This finding is consistent with what we observed in Figure~\ref{fig:simulation-normal1}, where the WASP approximation yielded a more smooth, regular posterior shape than the raw MCMC samples. Dependence on the number of subsets $K$ is modest once $K$ exceeds a small threshold. Overall, these results highlight WASP as a reasonable approximation to the oracle posterior, with tangible benefits over both naive subset posteriors and even full-data MCMC in terms of geometric regularity and covariance recovery.

\subsection{MNIST digits}

\begin{figure}[ht]
	\centering
	\includegraphics[width=.85\textwidth]{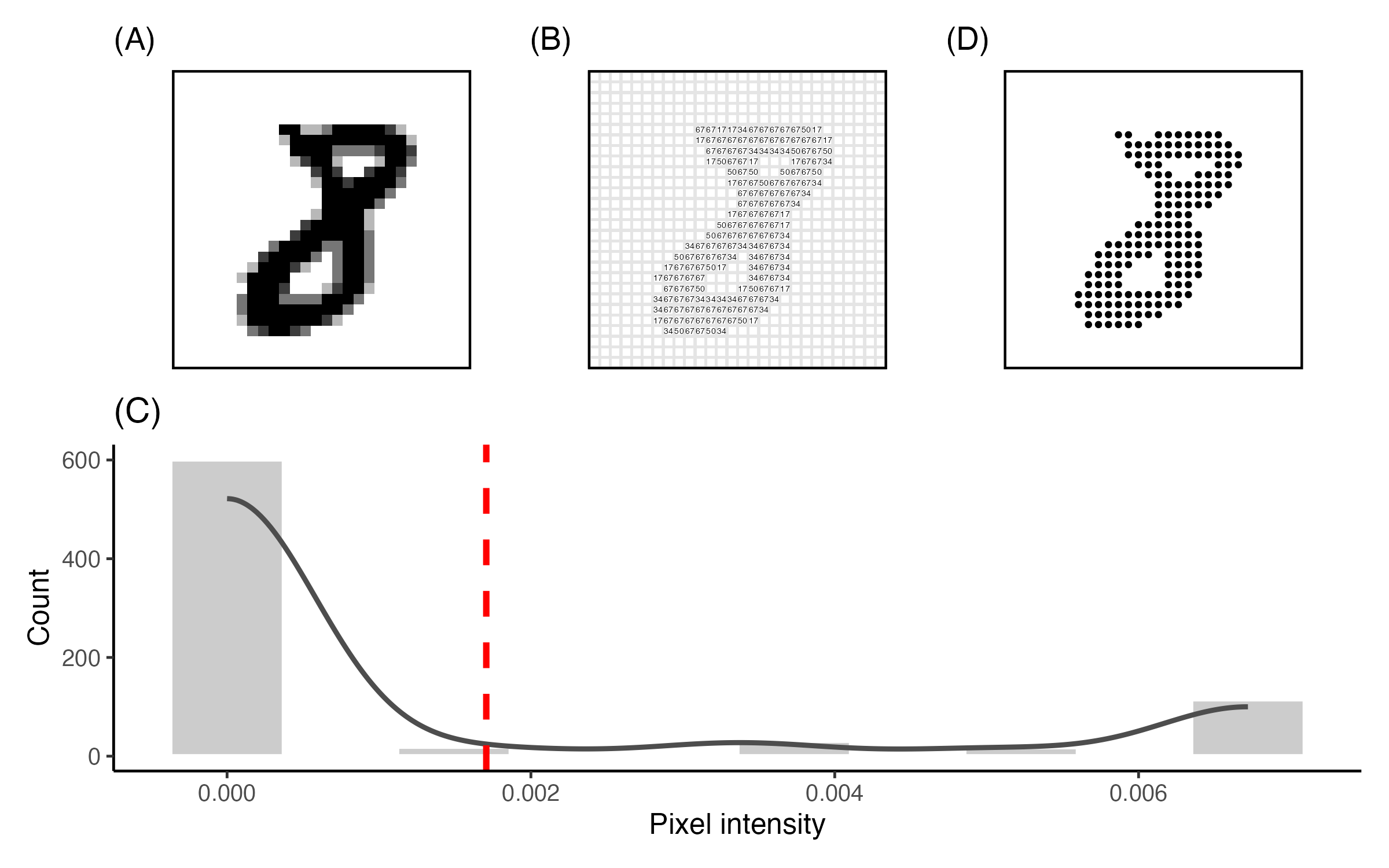}
	\caption{(A) Original $28\times 28$ grayscale raster of a digit ``8''. (B) Pixel lattice with intensities, where numbers are shown for illustration. (C) Empirical distribution of pixel intensities with kernel density overlay. The red dashed line marks the Otsu threshold. (D) Point-cloud representation after thresholding. Each dot is a foreground pixel and receives uniform weight.}
	\label{fig:real-digit1}
\end{figure}

Now we consider a real data setting using the MNIST handwritten digits \citep{lecun_1998_MNISTDatabaseHandwritten}. Each observation is a $28\times 28$ grayscale image with intensities in $[0,1]$. After $\ell_2$ normalization, an image may be interpreted as a probability distribution supported on the pixel lattice. For our particle-based barycenter algorithm, we instead view each image as an empirical measure in $\mathbb{R}^2$ whose support consists of the coordinates of active pixels. Concretely, we transform an image by thresholding its intensities through Otsu's method \citep{otsu_1979_ThresholdSelectionMethod} for threshold selection to obtain a binary mask of foreground pixels. The empirical measure then places uniform mass on the coordinates of those foreground pixels. This mapping suppresses low-amplitude background noise and yields sparser supports, which improves both visual interpretability and computational efficiency. Figure~\ref{fig:real-digit1} illustrates the conversion from raster to point cloud and the intensity distribution used to select the threshold. 

\begin{figure}[ht]
	\centering
	\includegraphics[width=.85\textwidth]{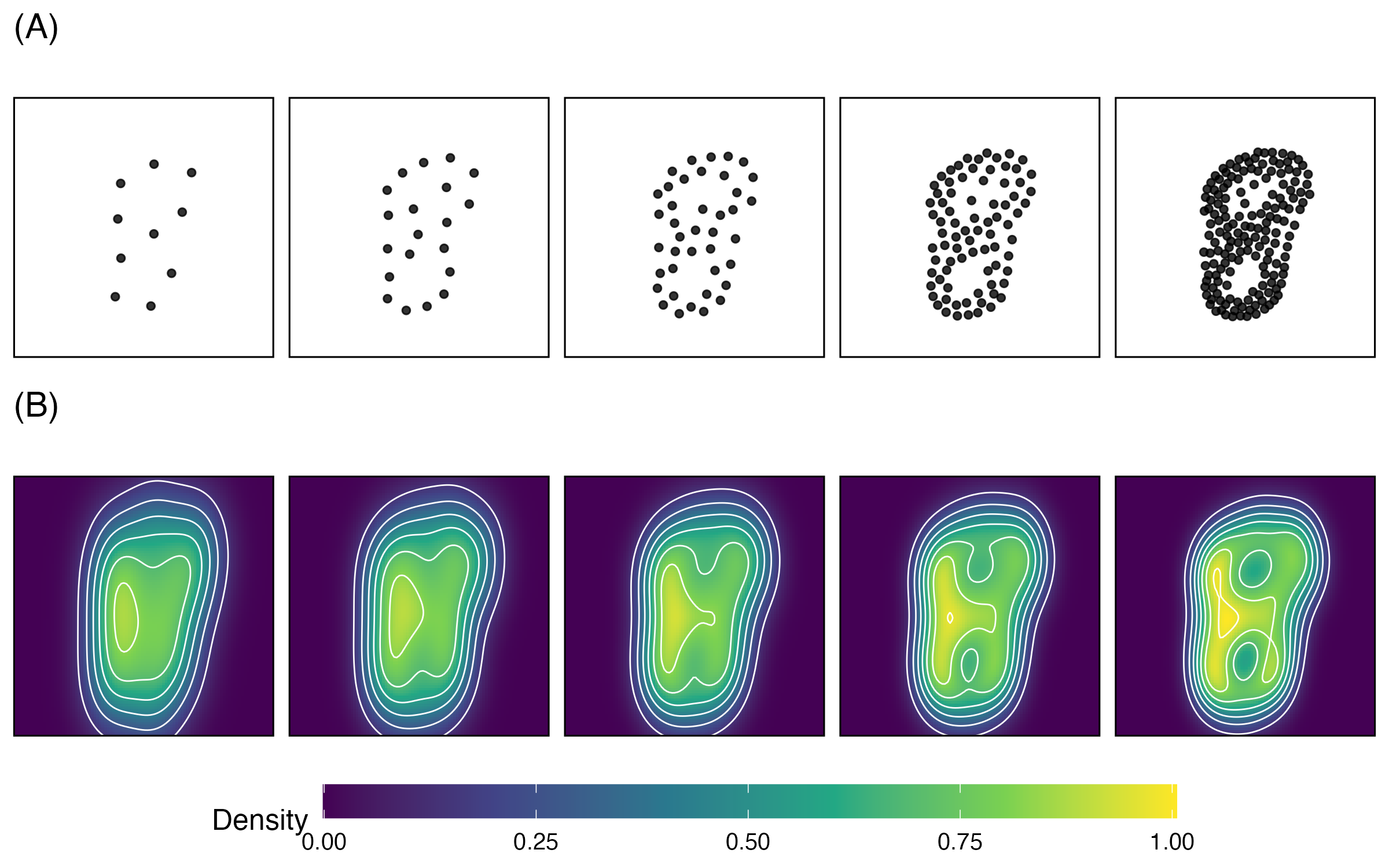}
	\caption{(A) Empirical barycenters of a digit ``8'' images with  cardinality $n \in \lbrace 10, 20, 40, 80, 160\rbrace$ from left to right. As $n$ increases, stroke geometric structure is resolved. (B) Corresponding kernel density estimates with overlaid contours. Internal holes and stroke thickness become well defined by $n = 80$, with finer structure visible at $n=160$.}
	\label{fig:real-digit2}
\end{figure}

Our goals in this experiment are twofold: (1) to examine how the barycenter's geometric and perceptual fidelity depends on its support size $n$, and (2) to assess the utility of class barycenters as prototypes for nearest-centroid classification under the $W_2$ metric. For each digit $c \in \lbrace 0, \ldots, 9\rbrace$, we randomly sample 100 training images, apply the aforementioned thresholding procedure that reduces the average cardinality of non-zero support by 28\% from 150.014 to 108.028, and compute a free-support barycenter with uniform barycenter weights and varying support sizes $n \in \lbrace 10, 20, \ldots, 200\rbrace$. Figure~\ref{fig:real-digit2} shows the evolution for a randomly selected digit ``8''. As $n$ increases, the particle configuration is expected to progressively resolve the character's strokes. Each barycenter goes through kernel density estimation to show how internal holes and stroke thickness are captured from a distributional perspective. 

Visual quality seems to improve monotonically with $n$ up to moderate resolutions. With $n=10$, the barycenter already sketches the digit's skeleton but can be visually ambiguous. At $n=40$ to $80$, the canonical shape of the two holes of an ``8'' is stably recovered. Beyond that, it further thickens strokes and begins to place particles in regions that reflect intra-class variability and some degree of occasional noise.

\begin{figure}[h]
	\centering
	\includegraphics[width=.7\textwidth]{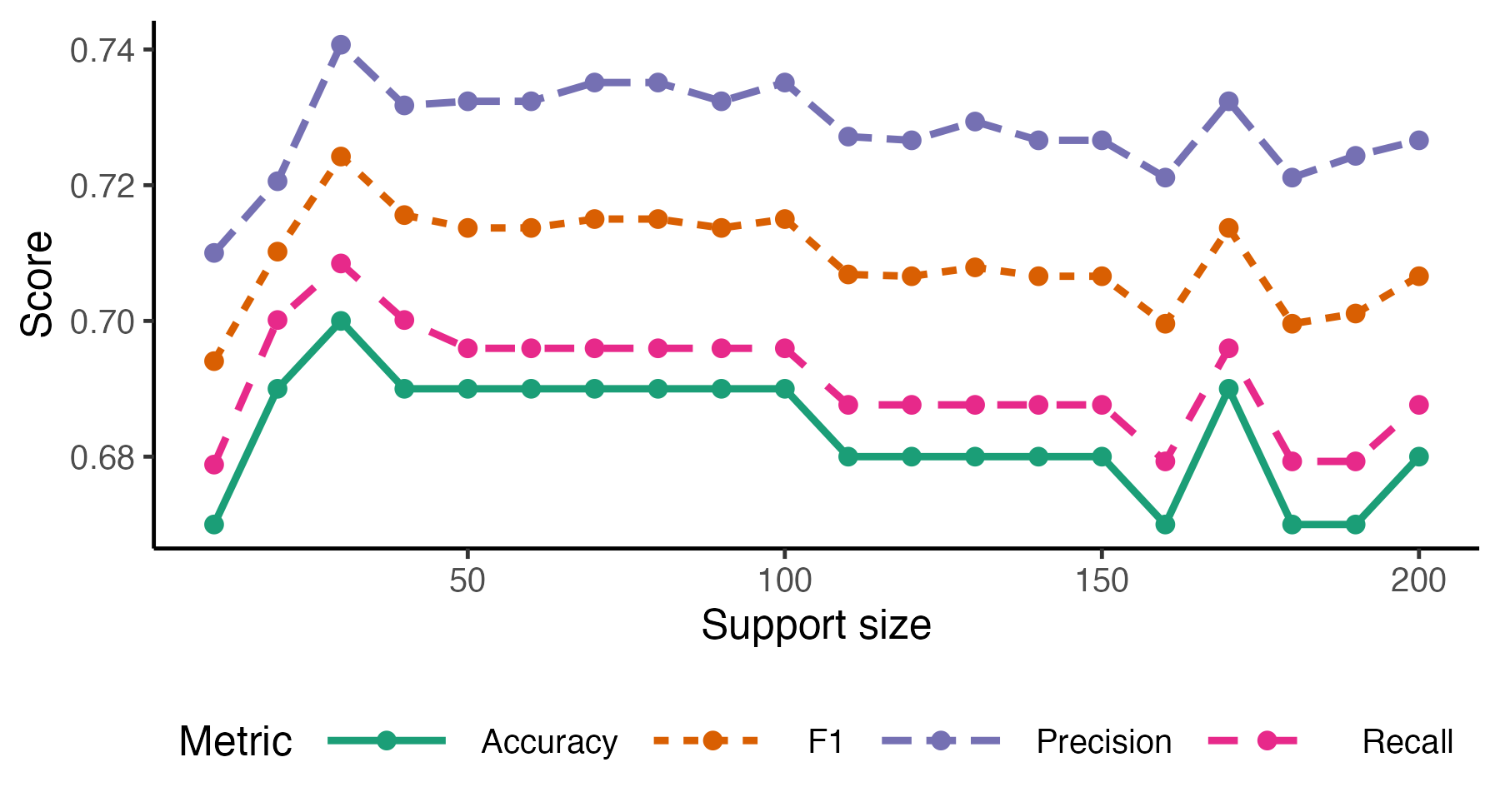}
	\caption{Accuracy, precision, recall, and F1 on a holdout set of 2,500 images as functions of varying support sizes $n \in \lbrace 10, 20, \ldots, 200\rbrace$.}
	\label{fig:real-digit3}
\end{figure}

For classification, we draw an independent set of 2,500 images with approximately balanced labels. Each test image is preprocessed in the same way and then assigned to the nearest class prototype with respect to the $W_2$ distance \citep{manning_2008_IntroductionInformationRetrieval}. We report accuracy, precision, recall, and F1 as functions of $n$  as shown in Figure~\ref{fig:real-digit3}. Classification exhibits a similar pattern of early gain and plateau. All metrics achieve their highest peak  near $n\approx 40$, remaining relatively stable thereafter. There is a modest bump visible around $n\approx 160$, likely reflecting better coverage of minor stroke patterns. Taken together, these curves indicate a practical operating regime of $n \in [40,80]$. In the range, the prototypical centroids are visually faithful and the classifier attains near-maximal performance without incurring the computation and mild overfitting tendencies observed at very large $n$. These findings are consistent with our simulation results that most of the benefit of increasing support size accrues quickly, after which residual error is dominated by within-class heterogeneity rather than discretization.

\subsection{Cluster analysis}

The last experiment is a large-scale clustering study that illustrates how the proposed barycenter computation can be used as building blocks for scalable partitional clustering. The standard $k$-means algorithm \citep{macqueen_1967_MethodsClassificationAnalysis} is a foundational baseline in cluster analysis with drawbacks in terms of memory- and time-demanding nature on large datasets. We therefore consider a distributed vector quantization (DVQ) scheme inspired by classical vector quantization \cite{gray_1998_Quantization}: (1) split the dataset into $S$ non-overlapping subsets, (2) compress each subset to a 10\% summary by running a $k$-means clustering with the corresponding number of centroids \citep{kreitmeier_2011_OptimalVectorQuantization}, and (3) cluster the union of subset summaries by running the proposed free-support barycenter solver with $k$ support points, which serve as DVQ centroids. This produces a $k$-prototype partition while touching only a fraction of the original points per iteration and allowing embarrassingly parallel computation for each subset. We compare DVQ against vanilla $k$-means and spherical $k$-means \citep{hornik_2012_SphericalMeansClustering} across a range of cluster sizes.

\begin{figure}[ht]
	\centering
	\includegraphics[width=.9\textwidth]{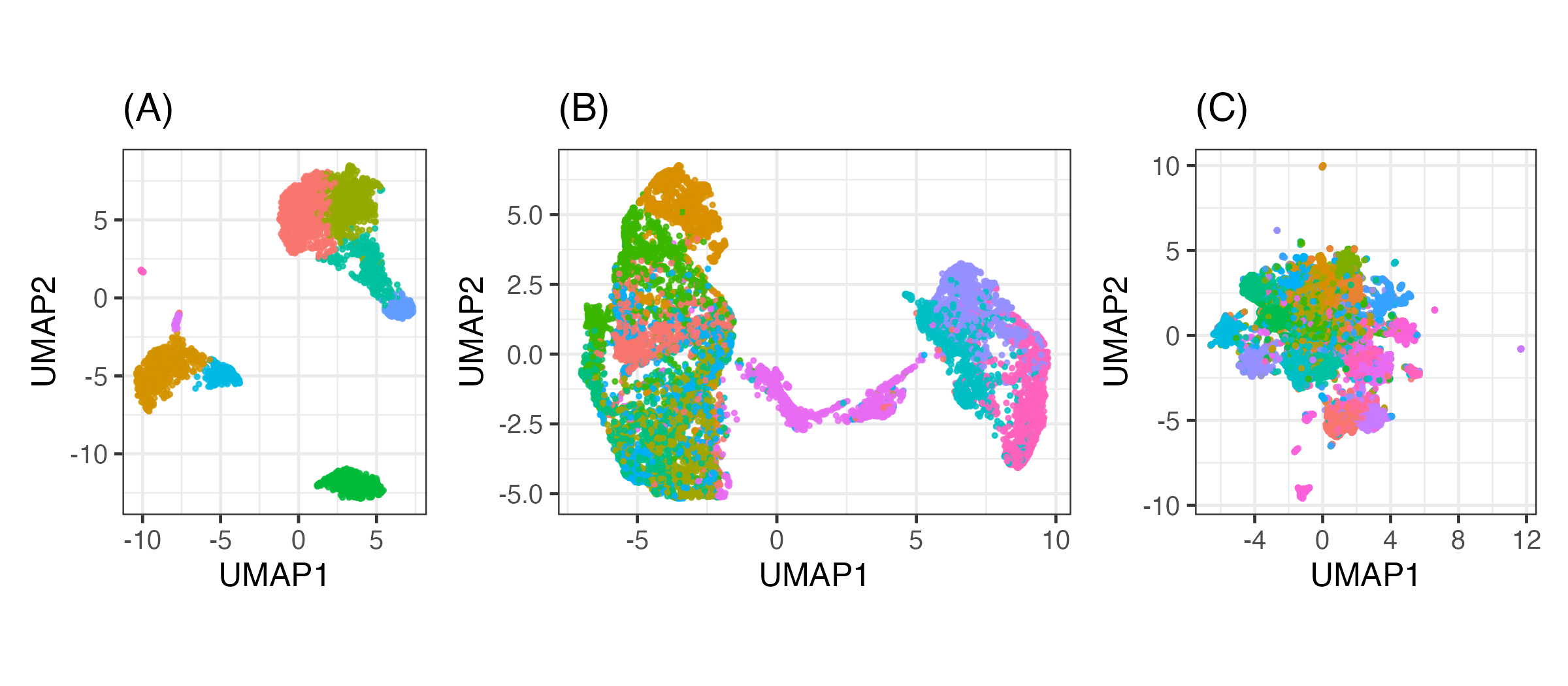}
	\caption{UMAP embeddings of the three datasets from the cluster analysis example in $\bbR^2$: (A) \textsf{PBMC}, (B) \textsf{NEWS}, and (C) \textsf{FASHION} data. Colors represent the ground-truth labels. }
	\label{fig:real-cluster1}
\end{figure}

We evaluate on three representative datasets. The first is \textsf{PBMC}, the single-cell peripheral blood mononuclear cells (PBMC) data that is publicly available on the 10x Genomics repository\footnote{\url{https://www.10xgenomics.com/datasets}}, processed using Seurat software \citep{hao_2024_DictionaryLearningIntegrative} by standard filtering, library-size normalization, log-transformation, and selection of the top 2,000 highly variable genes. The labels correspond to 9  cell types according to known canonical marker genes. The second is \textsf{NEWS}, the 20 Newsgroups collection texts  \citep{mitchell_1997_TwentyNewsgroups} of 20 topics represented as TF-IDF vectors in $\bbR^{24,164}$ after stop-word removal and document- and term-frequency filtering. The last is \textsf{FASHION}, the Fashion-MNIST test set with 10,000 grayscale images belonging to 10 classes. Each image matrix in $\mathbb{R}^{28\times 28}$ was vectorized to $\bbR^{784}$ after $\ell_2$ normalization. Figure~\ref{fig:real-cluster1} provides UMAP visualizations for each dataset in $\bbR^2$, colored by the ground-truth labels \citep{mcinnes_2020_UMAPUniformManifold}. For each dataset, we vary the pre-defined number of clusters $k$ across a standard range of $\pm 5$ from the true $k$, and report four commonly used criteria, including adjusted Rand index (ARI) \citep{rand_1971_ObjectiveCriteriaEvaluation}, normalized mutual information (NMI) \citep{strehl_2002_ClusterEnsemblesKnowledge}, average Silhouette score \citep{rousseeuw_1987_SilhouettesGraphicalAid}, and the Calinski-Harabasz (CH) index \citep{calinski_1974_DendriteMethodCluster}. DVQ results are shown for $S \in \lbrace 2, 5, 10\rbrace$ subsets.

\begin{figure}[ht]
	\centering
	\includegraphics[width=.9\textwidth]{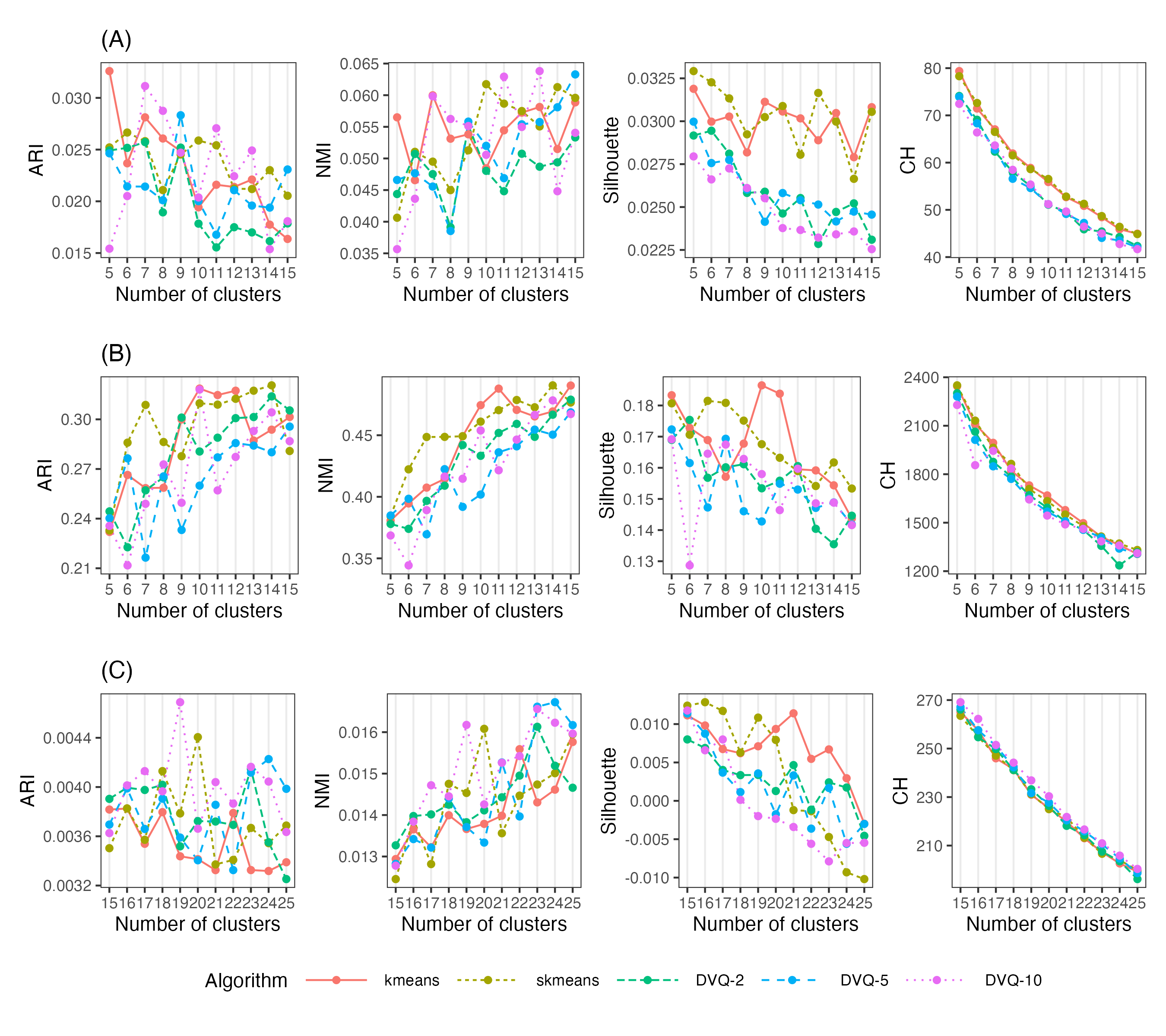}
	\caption{Clustering quality versus the number of clusters. Rows (A-C) correspond to \textsf{PBMC}, \textsf{NEWS}, and \textsf{FASHION} datasets, respectively. Columns show adjusted Rand index (ARI), normalized mutual information (NMI), Silhouette, and Calinski-Harabasz (CH) scores. The algorithms compared are $k$-means, spherical $k$-means, and DVQ with $S \in \lbrace 2, 5, 10\rbrace$ subsets.}
	\label{fig:real-cluster2}
\end{figure}

Results of the experiment are summarized in Figure~\ref{fig:real-cluster2}. We first recognize that these algorithms are performing almost equivalently across heterogeneous settings by considering the narrow range of vertical axes in all cases. On the \textsf{PBMC} dataset, DVQ with large $S$ attains the best ARI and NMI across most $k$, reflecting its ability to preserve fine cellular structure while denoising through subset-level quantization. Silhouette scores decrease mildly with $k$ as expected. When it comes to the \textsf{NEWS} dataset, DVQ with $S=5$ consistently matches spherical $k$-means in ARI and NMI, indicating that the barycentric prototypes remain relatively competitive in high-dimensional setting. A notable pattern is that improvements are most visible for moderate $k$, where topic granularity is well captured according to the ground truth. Lastly, all methods are close in ARI and NMI on the \textsf{FASHION} dataset with a definitive edge for DVQ-10 around the true number of clusters. Silhouette values are small for every method, which is potentially due to the overlapping nature of classes in raw pixel space. Another point of concern is that the CH reveals little in terms of all types of clustering patterns attained in our runs, which leads to questioning its efficacy in validating clustering structure in a highly nonlinear setting.  

Across all datasets, DVQ delivers compatible accuracy on par with the baselines while operating on approximately 10\% of the points per subset and enabling parallel solves. In short, free-support barycenters provide effective prototypes for large partitional clustering, offering a favorable accuracy-versus-efficiency trade-off without any entropic regularization.

\section{Conclusion}\label{sec:chp6-conclusion}

This paper develops a particle-flow perspective on free-support Wasserstein barycenters grounded in Otto’s formal Riemannian calculus. We show that barycenter computation can be viewed as first-order optimization on $\calP_2(\bbR^d)$, with updates driven by averaging of displacement fields. In discrete settings where Monge maps are unavailable, we replace them with barycentric projections of optimal transport plans, obtaining an unregularized and geometry-faithful algorithm that preserves sharp structures and avoids the smoothing artifacts endemic to entropic regularization.

Our contribution includes theoretical results that clarify when and why this surrogate is principled. We prove that barycentric projections consistently approximate the logarithmic map as empirical measures densify, that the resulting scheme enjoys monotone descent and converges to stationary points under natural local stability of the plans, that the objective is stable with respect to perturbations of the inputs, and that increasing the number of barycenter atoms yields resolution consistency toward the population barycenter. Collectively, these statements establish that the discrete algorithm tracks the continuous gradient flow to first order and that accuracy improves predictably with representational capacity. Through numerical experiments, we provide empirical grounds to support the effectiveness of the proposed framework in a number of practical tasks, including averaging continuous probability distributions from the semi-discrete approach, scalable Bayesian computation by aggregating subset posteriors, image representation and measure-based classification, and large-scale clustering. All of the experiments benefit from the characteristic of the proposed algorithm to freely vary the resolution of the target empirical barycenter.

These benefits, however, are not without limitations. The Riemannian structure we employ is formal. While it reliably guides first-order methods, classical differential-geometric tools are generally unavailable. Nonconvexity in the free-support objective implies convergence to stationary points rather than global optima, making initialization important in practice. Barycentric projections introduce a surrogate bias at finite resolution, even though it vanishes asymptotically. Despite intrinsic parallelism, each iteration still requires solving multiple OT problems at each iteration, which remains the dominant cost. We also fixed barycenter weights for clarity. Joint optimization of support locations and weights may improve fidelity, but it may complicate both analysis and implementation.

The findings open several promising directions for future work. On the theoretical side, non-asymptotic bounds that disentangle discretization, projection, and sampling errors would provide practical guidance for choosing decent support sizes that balance the computational efficiency and representational effectiveness. It is also interesting to characterize conditions under which the local geometry near the barycenter yields faster rates under mild regularity. Algorithmically, acceleration schemes, such as momentum updates adapted to transport, adaptive step rules, and variance-reduced variants for federated settings, merit exploration. We have already observed practical gains from multiscale refinement of supports, and we invite formal complexity analyses. An equally important direction is to broaden the modeling envelope. One direct avenue is to apply the particle-flow approach to computation of the Wasserstein median \citep{you_2025_WassersteinMedianProbability}, which could eliminate the nested use of barycenter computation within the iteratively reweighted least squares framework. It is also of interest to extend the method to non-Euclidean supports, such as Riemannian manifolds, to enable measure-valued data analysis in the empirical regime. In parallel, there is an opportunity to make barycenters a differentiable building block for modern learning systems: unrolling a finite number of our updates yields a barycenter layer for end-to-end training as a distributional pooling or prototype module. Finally, robust variants including trimmed flows, median-of-measures schemes, or adversarially regularized displacements could improve resilience to outliers and model misspecification common in scientific data.

\bibliographystyle{dcuky}
\bibliography{references}


\end{document}